\documentclass{article}

\PassOptionsToPackage{dvipsnames,table}{xcolor}

\ifdefined\pdfsuppressptexinfo\pdfsuppressptexinfo=-1\fi

\usepackage{iclr2027_conference,times}

\usepackage{amsmath,amsfonts,bm}

\def\eqref#1{equation~\ref{#1}}

\def\1{\bm{1}}

\def\vb{{\bm{b}}}
\def\vc{{\bm{c}}}

\def\vh{{\bm{h}}}

\def\vu{{\bm{u}}}

\def\vx{{\bm{x}}}

\def\vz{{\bm{z}}}

\def\mW{{\bm{W}}}

\DeclareMathAlphabet{\mathsfit}{\encodingdefault}{\sfdefault}{m}{sl}
\SetMathAlphabet{\mathsfit}{bold}{\encodingdefault}{\sfdefault}{bx}{n}

\newcommand{\E}{\mathbb{E}}

\usepackage{hyperref}
\usepackage{url}

\newcommand{\anisoflowbibkey}{anonymous2026anisoflow}
\AtBeginDocument{%
  \ificlrfinal\renewcommand{\anisoflowbibkey}{zirvi2026anisoflow}\fi
  \ifarxiv\renewcommand{\anisoflowbibkey}{zirvi2026anisoflow}\fi
}

\usepackage{amssymb}
\usepackage{mathtools}
\usepackage{amsthm}
\usepackage{graphicx}
\usepackage{booktabs}
\usepackage{caption}
\usepackage{enumitem}
\usepackage[dvipsnames]{xcolor}

\usepackage{float}
\usepackage{wrapfig}
\usepackage{multirow}
\usepackage{makecell}
\usepackage[linesnumbered,ruled,vlined]{algorithm2e}
\usepackage{etoolbox}

\makeatletter
\patchcmd{\@algocf@start}{-1.5em}{-0.5em}{}{}
\makeatother

\SetCommentSty{mycommfont}
\colorlet{lnum}{black!60}

\SetNlSty{nlnumstyle}{}{}
\SetAlgoNlRelativeSize{-1}
\definecolor{teacher}{RGB}{78,167,46}
\definecolor{student}{RGB}{21,96,130}
\definecolor{detached}{RGB}{140,140,140}

\newlength\itemparskip
\setlist{nosep, leftmargin=1em, itemsep=0pt, parsep=0pt, before={\parskip=\itemparskip}, after=\vspace{-\itemparskip}}%

\newtheoremstyle{custom}
{\parskip} %
{0}
{} %
{} %
{\bfseries} %
{.} %
{.5em} %
{} %
\theoremstyle{custom}

\theoremstyle{definition}

\theoremstyle{remark}

\definecolor{basecol}{HTML}{C060C0}
\definecolor{dwdcol}{HTML}{5CB85C}
\newcommand{\qualhead}{%
  \noindent\makebox[0.49\linewidth][c]{\large\bfseries\textcolor{basecol}{Baseline}}\hfill%
  \makebox[0.49\linewidth][c]{\large\bfseries\textcolor{dwdcol}{DWD}}\par\vspace{2pt}}
\newcommand{\qualpair}[3]{%
  {\setlength{\fboxsep}{0pt}\setlength{\fboxrule}{1pt}%
   \noindent\fcolorbox{basecol}{white}{\includegraphics[width=0.485\linewidth]{figures/app_qual/#1_baseline.jpg}}\hfill%
   \fcolorbox{dwdcol}{white}{\includegraphics[width=0.485\linewidth]{figures/app_qual/#1_dwd.jpg}}\par}%
  \vspace{1pt}{\footnotesize\itshape\noindent #2}\par\vspace{#3}}

\iclrfinalcopy
\arxiv

\begin{document}

\title{Decoding Without a Decoder: Direct Pixel Prediction from Diffusion Transformers}

\author{%
  \vspace{0ex}
  \textbf{Hansheng Chen}$^{1,2}$ \quad 
  \textbf{Shengqu Cai}$^{1}$ \quad 
  \textbf{Jan Ackermann}$^{1}$ \quad 
  \textbf{Rayhan Zirvi}$^{1,3}$ \\
  \vspace{1ex}
  \textbf{Gordon Wetzstein}$^{1}$ \quad 
  \textbf{Xinchen Yan}\thanks{Work done while at Luma AI.} \quad 
  \textbf{Leonidas Guibas}$^{1}$ \\
  \vspace{1ex}
  $^{1}$Stanford University \quad
  $^{2}$Luma AI \quad
  $^{3}$California Institute of Technology \\
  \vspace{-1ex}
  \url{https://hanshengchen.com/dwd}
}

\maketitle

\begin{figure}[H]
\centering
\includegraphics[width=\linewidth]{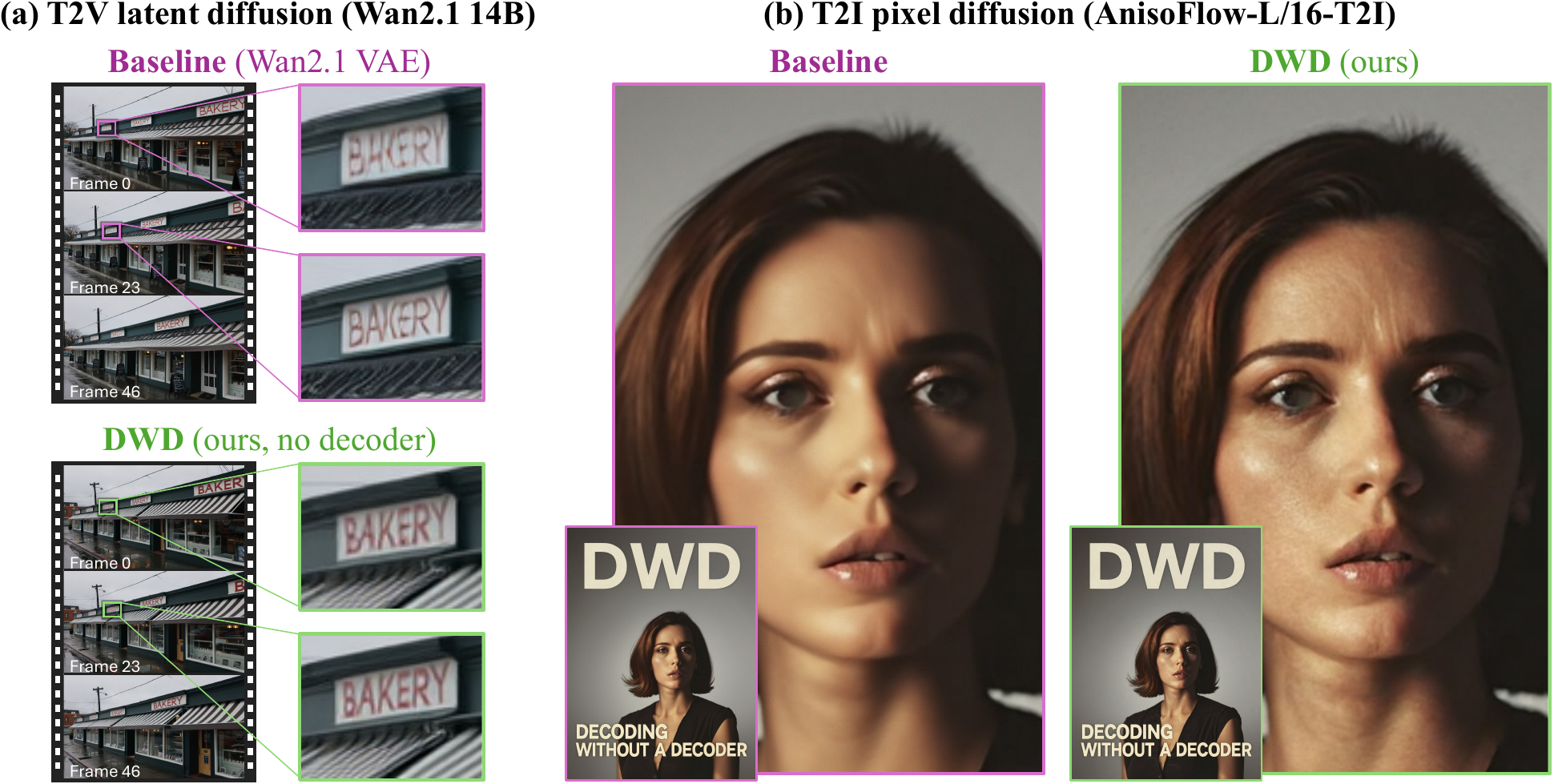}
\caption{\textbf{Given the same seed, DWD preserves similar content to the original model while improving fine detail.}  \textbf{(a)} On latent-space video diffusion, DWD removes the decoder and keeps text rendering legible across frames, whereas the baseline distorts it. \textbf{(b)} On pixel-space image diffusion, DWD recovers photorealistic skin and hair texture that the baseline oversmooths.}
\label{fig:teaser}
\end{figure}

\begin{abstract}
Latent diffusion models rely on a decoder to map denoised latents to pixels, adding complexity and inference cost. We introduce \emph{Decoding Without a Decoder} (DWD), which lets a finetuned diffusion transformer produce final pixels directly through a linear head. Unlike previous methods, where the denoising output leads to final pixels, DWD separates the two roles: the original denoising head continues to drive sampling, then the new DWD head predicts the pixels at the final step. We extend pretrained latent models with DWD efficiently by finetuning only the last few transformer blocks, training the DWD head with pixel quality losses while anchoring the denoising head to the pretrained model to preserve the sample trajectory. Across latent image and video generators, DWD removes the decoder while maintaining or even improving pixel quality. On Wan video generation, it visibly reduces detail distortion and flickering, lowering warping error by 23\%. Remarkably, the same separation also improves pixel diffusion models that do not even use a decoder: a final-step DWD head produces substantially better texture detail in text-to-image generation, and improves ImageNet FID from 1.55 to 1.43, a new state of the art among pixel flow models.\ifarxiv\else{} Code and models will be released.\fi
\end{abstract}

\section{Introduction}
\label{sec:intro}

Latent diffusion~\citep{rombach2022ldm,ddpm,song2021scorebased,lipman2023flow} is the dominant recipe for image and video generation~\citep{esser2024sd3,flux2024,wan,hunyuanvideo,ltxvideo,zimage,flux2}. Typically, a diffusion transformer (DiT)~\citep{dit} generates in a compressed latent space, and a convolutional decoder~\citep{kingma2014vae,esser2021vqgan} maps the latents to pixels. The decoder complicates deployment and can add substantial inference cost. For example, on H100, generating a 5-second video with a 4-step Wan 1.3B model~\citep{rcm} takes 2.8 seconds for sampling and another 1.9 seconds for decoding, adding about 70\% overhead. Recent work obtains much faster decoders with comparable quality~\citep{zhu2026flashvaed,kang2026flashdecoder,li2026solh3,gu2026lightvae}, suggesting that the decoder is largely redundant. This raises the question: what is the minimal network architecture for producing final pixels from a latent diffusion model?

We show that a simple linear head can replace the decoder while improving pixel quality. Our method, \emph{Decoding Without a Decoder} (DWD), finetunes a pretrained latent DiT by adding a linear output head (the \emph{DWD head}) in parallel with the DiT's original output layer (the \emph{denoising head}). The denoising head continues to drive denoising steps, then the DWD head predicts the pixels at the final step only.

The dual-head separation distinguishes DWD from previous approaches, whose denoising output leads to final pixels: latent models decode it, pixel models output it directly. This couples pixel fidelity to denoising, causing a tradeoff between detail rendering and denoiser learnability~\citep{yao2025vavae,skorokhodov2025diffusability}. DWD decouples the two roles: the DWD head is trained with pixel quality losses~\citep{esser2021vqgan,lpips}, while the denoising head is anchored to the frozen pretrained model to preserve sample content.

The separation is most evident on pixel diffusion models, where the denoising head already predicts the final output. These predictions smooth fine detail because they estimate conditional means under $\ell_2$ loss. Applying perceptual losses directly to them~\citep{lin2023perceptual,berrada2025boosting,ma2026pixelgen} risks collapsing sample diversity, as our shared-head ablation fails to improve FID (Table~\ref{tab:latent}). In contrast, a separate head improves ImageNet FID from 1.55 to 1.43.

We evaluate DWD across latent and pixel generators on ImageNet, text-to-image, and video generation (Fig.~\ref{fig:teaser}). Removing the decoder improves nearly all FID metrics across tested latent image models. On Wan~\citep{wan} video, warping error drops by 23\%, with visibly less detail distortion and flickering. On text-to-image pixel diffusion, DWD recovers photorealistic textures.

To summarize, we show that:
\begin{itemize}
    \item A latent DiT needs no decoder: finetuning its last few blocks with an additional linear head suffices, reducing inference cost, improving ImageNet FIDs, and lessening detail distortion and flickering in Wan video.
    \item The dual-head separation also improves pixel diffusion models, producing better texture detail in text-to-image generation and a state-of-the-art ImageNet FID of 1.43 among pixel flow models.
\end{itemize}

\section{Related Work}
\label{sec:related}

\textbf{Latent-space diffusion.}
VQGAN~\citep{esser2021vqgan} trains autoencoders with reconstruction, perceptual~\citep{lpips}, and adversarial~\citep{goodfellow2014generative} losses, and LDM~\citep{rombach2022ldm} uses this perceptual compression to reduce diffusion cost.
VA-VAE~\citep{yao2025vavae}, REPA-E~\citep{leng2025repae}, and RAE~\citep{zheng2025rae} address the gap between reconstruction quality and diffusability.
DC-AE~\citep{chen2024dcae} and Qwen-Image-VAE-2.0~\citep{qwenvae2} increase spatial compression while preserving reconstruction fidelity.
SSDD~\citep{vallaeys2025ssdd} and PiD~\citep{lu2026pid} replace conventional decoders with distilled pixel diffusion models. PiD and LTX-Video~\citep{ltxvideo} also stop latent sampling early, letting the decoder complete the remaining denoising.
\citet{zhang2024pixelpost} improve latent generators by applying losses to images produced through the VAE decoder.
These approaches retain a standalone decoding network.
DWD instead produces the final pixels through a linear head on the existing DiT, reusing its last-layer tokens and computation without an extra decoder.

\textbf{Pixel-space diffusion.}
Pixel-space diffusion models denoise directly in pixel space without an autoencoder. They overcome difficulties in high-resolution generation through multiscale architectures and specialized output pathways~\citep{hoogeboom2023simple,crowson2024scalable,yu2025pixeldit,wang2025pixnerd,ma2025deco}, or improved prediction targets~\citep{li2025jit,chen2026asymflow,\anisoflowbibkey}.
AsymFlow~\citep{chen2026asymflow} and L2P~\citep{chen2026l2p} finetune pretrained latent models to perform denoising in pixel space, altering the flow trajectory and sample content (see Appendix~\ref{app:asymflow} for a comparison). In contrast, DWD preserves the original latent trajectory rather than converting the model to pixel diffusion. DWD finetuning is also more efficient, since only the last few transformer blocks and output heads are updated with shallow backpropagation, whereas latent-to-pixel conversion must backpropagate through the entire network to adapt to pixel-space inputs.

\section{Decoding Without a Decoder}
\label{sec:method}

\begin{figure}[t]
\centering
\includegraphics[width=\linewidth]{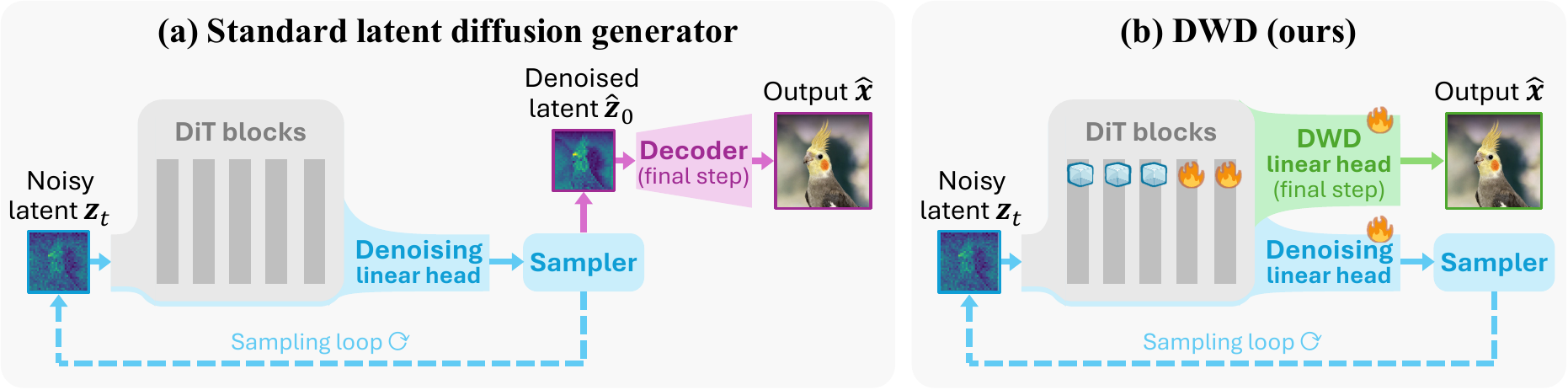}
\caption{\textbf{Standard latent generation vs.\ DWD.} \textbf{(a)} A standard generator samples a denoised latent and maps it to pixels with a decoder. \textbf{(b)} DWD retains the denoising head for latent denoising and adds a parallel DWD head that produces the final pixels only at the final step, bypassing this decoder. DWD finetunes the pretrained generator's final blocks and heads (fire icon), while freezing all previous blocks (ice icon). Time $t$ and conditioning $\vc$ are omitted for clarity.}
\label{fig:method}
\end{figure}

A pretrained latent DiT's last-layer tokens already encode rich pixel information~\citep{gan2025ditf,akarken2026mmdiff,wang2026rechannel}: a least-squares linear projection recovers most of the image structure before any finetuning (Fig.~\ref{fig:probe}). DWD exploits this by appending a linear head that produces final pixels directly from these tokens, bypassing the decoder at inference (Fig.~\ref{fig:method}(b)). Separating the pixel readout from the denoising output lets each be trained with its own objective: pixel quality losses for the DWD head, and trajectory anchoring to preserve denoising behavior.

\subsection{Preliminaries}
\label{sec:prelim}

We describe latent diffusion~\citep{rombach2022ldm} using the flow matching convention~\citep{lipman2023flow}. A pretrained autoencoder encodes an image or video $\vx$ into a latent representation $\vz_0$. The forward process mixes the latent $\vz_0$ with Gaussian noise to form $\vz_t=(1-t)\vz_0+t\bm{\epsilon}$, where $\bm{\epsilon}\sim\mathcal{N}(\mathbf{0},\mathbf{I})$ and $t\in[0,1]$ denotes diffusion time. A flow model $(\vz_t,t,\vc) \mapsto \hat{\vu}$ trains a network with extra conditioning $\vc$ to estimate the conditional mean of velocity $\vu \coloneqq \bm{\epsilon} - \vz_0$, using an $\ell_2$ flow matching loss. At inference, the model uses this velocity to progressively transform Gaussian noise into a denoised latent sample $\hat{\vz}_0$, running from $t=1$ toward $t=0$~\citep{song2021scorebased,liu2022rectifiedflowmarginalpreserving,albergo2023building}. Finally, the decoder maps $\hat{\vz}_0$ to the final image or video $\vx$.

For the models studied here, the denoising network is a diffusion transformer (DiT)~\citep{dit,esser2024sd3}, which processes tokenized noisy latent $\vz_t$ with multiple transformer blocks~\citep{attention}, and feeds the last-layer tokens $\vh$ (after normalization and modulation) to a linear output head that produces a patch-wise denoising prediction. This is the \emph{denoising head}, shown in Fig.~\ref{fig:method}(a).

\subsection{DWD head}
\label{sec:method_head}

Our goal is to produce final pixels from a pretrained latent DiT without a decoder, subject to two constraints: minimal additional computation, and preservation of the pretrained denoising trajectory.

\begin{figure}[t]
\centering
\includegraphics[width=\linewidth]{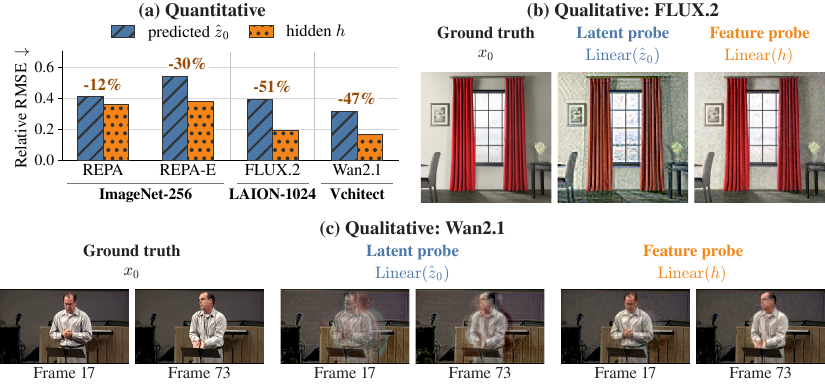}
\vspace{-3.3ex}
\caption{
\textbf{Last-layer tokens support more accurate linear pixel reconstruction.} We fit patch-wise linear probes from denoised latents $\hat \vz_0$ and last-layer tokens $\vh$ to ground-truth pixels, keeping the pretrained models frozen. \textbf{(a)} Held-out relative RMSE, with percentages showing error reduction over the latent probe. \textbf{(b, c)} Probes of last-layer tokens preserve finer image detail and reduce video ghosting before finetuning.
}
\label{fig:probe}
\end{figure}

A linear probe shows that the last-layer tokens $\vh$ already support pixel reconstruction without finetuning, more accurately than probing the denoised latents $\hat{\vz}_0$, as shown in Fig.~\ref{fig:probe}. The remaining reconstruction mismatch can be addressed by finetuning the DiT. This inspires us to adopt a simple linear \emph{DWD head} that maps each token to a pixel patch:
\begin{equation}
    \hat{\vx}_i = \operatorname{Linear}(\vh_i) = \mW\vh_i + \vb,
    \label{eq:pixel_head}
\end{equation}
where $\mW$ and $\vb$ are learnable weights and bias, and $\hat{\vx}_i$ is the pixel patch corresponding to token $\vh_i$. This design places the DWD head in parallel with the denoising head, both reading the same last-layer tokens. The linear head is the minimal form of this separate pixel output, which can optionally be extended with a lightweight convolutional refiner (Appendix~\ref{app:head}). The predicted patches are then assembled into the final image or video $\hat{\vx}$, with color space conversion where needed (Appendix~\ref{app:implementation}).

\textbf{Efficiency.} 
By reusing the DiT's nonlinear computation, the linear head is a lightweight readout with negligible cost for common pixel patch sizes. Pixel patch size follows the combined downsampling factor of the autoencoder and tokenizer. The most common spatial factor of 16 gives a 16\texttimes16\texttimes3 patch per token. For video models, an additional temporal factor of 4 expands each patch to four frames. For most tested models, the total number of output dimensions per token stays below the DiT's hidden dimension.

\textbf{Separating denoising and pixel outputs.}
DWD achieves trajectory preservation and pixel rendering through a clear separation of outputs: the DWD head is trained with pixel quality losses, while the denoising head drives sampling and is anchored to the pretrained model (Sec.~\ref{sec:method_training}). This dual-head separation distinguishes our work from prior diffusion-based image and video generators, where the denoising output serves as a bottleneck pathway to final pixels.

\textbf{Final-step pixel rendering.}
The DWD head is only used at the final denoiser evaluation step, where the DiT input $t$ is small but nonzero. The nonzero timestep is important: treating DiT--DWD as a latent decoder at $t{=}0$ substantially worsens quality (Sec.~\ref{sec:exp_terminal}), while the $t{>}0$ noise gives the pixel head freedom to create detail that the denoised latent would have smoothed away. Classifier-free guidance~\citep{cfg} is not used in the final step since its role is to steer the trajectory, whereas our final step does not advance the trajectory.

\subsection{Finetuning}
\label{sec:method_training}

We finetune the last $K$ transformer blocks (typically $K{=}4$--$8$ for large text-conditioned models, see Appendix~\ref{app:implementation} for details), both output heads, and the last modulation, keeping the prefix frozen. The timestep embedding MLP is shared across all blocks, so we duplicate its weights for the adapted blocks, allowing their modulations to be fully tuned without disturbing the frozen prefix. This restricts backpropagation to the end of the network and retains most pretrained parameters, making training up to 3\texttimes\ faster than full backpropagation.

The training objective combines pixel quality with trajectory anchoring:
\begin{equation}
    \mathcal{L} = \mathcal{L}_{\mathrm{pix}} + \lambda\mathcal{L}_{\mathrm{traj}},
    \label{eq:total_loss}
\end{equation}
where $\lambda$ is the trajectory anchoring weight (Appendix~\ref{app:implementation}). Both terms use noisy inputs $\vz_t$ encoded from real image data $\vx$, but their timesteps $t$ are sampled separately to match the roles of the two outputs. The two losses are backpropagated before one optimizer step on accumulated gradients, so the last-layer tokens adapt for pixel quality while the denoising head's prediction remains anchored to the pretrained model. The full training loop is shown in Algorithm~\ref{alg:dwd_training}.

\ifarxiv
\begin{figure}[!b]
\centering
\begin{minipage}{0.54\textwidth}
\input{algorithms/dwd_algorithm}
\end{minipage}
\end{figure}
\fi

\textbf{Pixel quality.}
The DWD head predicts the clean image under the L1 reconstruction, LPIPS perceptual, and adversarial objectives commonly used to train autoencoders~\citep{esser2021vqgan,rombach2022ldm,lpips,goodfellow2014generative}:
\begin{equation}
    \mathcal{L}_{\mathrm{pix}} = \E_{\vx,\vz,\vc,t\sim p_{\mathrm{pix}}}\Big[\|\hat{\vx}-\vx\|_1 + \mathrm{LPIPS}(\hat{\vx},\vx) + \ell_{\mathrm{adv}}(\hat{\vx})\Big],
    \label{eq:pixel_loss}
\end{equation}
where $p_{\mathrm{pix}}$ is typically uniform over $[0.0, 0.1]$ before timestep shifting~\citep{esser2024sd3}, matching the small-$t$ regime where the DWD head operates at inference. Following \citet{zheng2025rae} and \citet{raev2}, the adversarial term uses the StyleGAN-T discriminator~\citep{sauer2023stylegant} with adaptive gradient balancing, and is switched on only in the last stage of training (Appendix~\ref{app:implementation}).

\textbf{Trajectory anchoring.}
To preserve the pretrained denoising behavior as the final blocks adapt, we match the denoising head's prediction to that of a frozen copy of the original model. Let $\hat{\vu}$ and $\hat{\vu}^{\mathrm{ref}}$ denote the adapted and frozen reference velocity predictions for the same input. We minimize
\begin{equation}
    \mathcal{L}_{\mathrm{traj}} = \E_{\vx,\vz,\vc,t\sim p_{\mathrm{traj}}}\Big[\big\|\hat{\vu}-\hat{\vu}^{\mathrm{ref}}\big\|_2^2\Big] ,
    \label{eq:reg_loss}
\end{equation}
where $p_{\mathrm{traj}}$ is typically uniform over $[0.04, 1.0]$ before timestep shifting, covering the intermediate denoising steps. Both the original and finetuned models share a single forward pass through the frozen prefix: only the final $K$ blocks and output components run separately, avoiding two full-model passes.

\ifarxiv\else
\begin{wrapfigure}{r}{0.54\textwidth}
\vspace{-0.5\intextsep}
\input{algorithms/dwd_algorithm}
\vspace{-2\intextsep}
\end{wrapfigure}
\fi

\subsection{Extension to pixel diffusion and few-step models}
\label{sec:method_ext}

\textbf{Pixel diffusion models.}
The quality benefit of the dual-head separation is most pronounced on pixel diffusion models, which have no decoder to confound the comparison.
We still add a DWD head in parallel with the denoising head, treating pixel denoising the same as the latent case. The training objectives are unchanged except for a stronger anchoring weight. Fewer tunable blocks and a shorter training schedule suffice, since the DiT already has pixel knowledge. Details are in Appendix~\ref{app:implementation}.

\textbf{Few-step latent models.}
Few-step distilled models~\citep{yin2024onestep,yin2024improved,ladd,chen2025piflow} are the practical frontier of fast generation, where the fixed decoding cost takes a proportionally larger share of inference time. Validating DWD here therefore has direct practical significance. A simple adaptation is to adjust $p_{\mathrm{pix}}$ and $p_{\mathrm{traj}}$ to cover new ranges around the timesteps used by the few-step sampler. \ifarxiv More variants are discussed in Appendix~\ref{app:fewstep_video} and \ref{app:mmh3}.\else Another variant is discussed in Appendix~\ref{app:fewstep_video}.\fi

\section{Experiments}
\label{sec:exp}

\subsection{ImageNet generation}
\label{sec:exp_imagenet}

We show that DWD consistently improves FID across both latent and pixel baselines. We then ablate the separate head design and show that fusing the two heads degrades quality despite the same training losses, and also study the effect of head capacity and final timestep.

\textbf{Setup.}
We test two latent models, SiT-XL/2-REPA~\citep{ma2024sit,yu2024repa} with the SD-VAE decoder~\citep{rombach2022ldm} and SiT-XL/2-REPA-E~\citep{leng2025repae} with its end-to-end tuned E2E-VA-VAE decoder~\citep{yao2025vavae}, and two pixel models, JiT-H/16~\citep{li2025jit} and AnisoFlow-H/16~\citep{\anisoflowbibkey}. DWD extensions are finetuned on ImageNet~\citep{imagenet}, with training schedules detailed in Appendix~\ref{app:finetuning}. We report FID~\citep{FID} and Inception Score (IS)~\citep{salimans2016is} under the ADM evaluation protocol~\citep{dhariwal2021adm}.

\begin{table}[t]
\centering
\caption{\textbf{Results on ImageNet 256\texttimes 256.} {\setlength{\fboxsep}{1pt}\colorbox{blue!6}{Shaded rows}} are our DWD extensions, finetuned from the baseline above them. Italic rows are ablations: \emph{DWD-conv} adds convolutional refinement, and \emph{w/o separate head} fuses the DWD and denoising heads so that a single head receives both pixel losses and trajectory anchoring. DWD has slightly more parameters than the baseline denoiser mainly due to the duplicated timestep embedding MLP.}
\label{tab:latent}
\scalebox{0.9}{%
\begin{tabular}{@{}c@{\hspace{8pt}}>{\columncolor{white}[0pt][0pt]}p{1em}@{}>{\columncolor{white}[0pt][\tabcolsep]}lccccc}
\toprule
& \multicolumn{2}{@{}l}{\textbf{Method}} & \textbf{Decoder} & \small\makecell[c]{\textbf{Params (M)} \\ (+\,decoder)} & \small\makecell[c]{\textbf{GFLOPs} \\ (final step)} & \textbf{FID\textdownarrow} & \textbf{IS\textuparrow} \\
\midrule
& \multicolumn{2}{@{}l}{SiT-XL/2-REPA} & SD-VAE & 674.8\,+\,49.5 & \textcolor{gray!70}{2\,\texttimes}\,237.2\,+\,622.2 & 1.38 \hphantom{(+0.00)} & 306.2 \hphantom{(+0.0)} \\
\rowcolor{blue!6} \cellcolor{white} & & \textbf{DWD} & -- & 678.5\hphantom{\,+\,00.0} & \hphantom{2\,\texttimes\,}237.7\hphantom{\,+\,000.0} & 1.36 \textcolor{Green}{(\textminus0.02)} & 312.5 \textcolor{Green}{(+6.3)} \\
& & \emph{DWD-conv} & -- & 678.9\hphantom{\,+\,00.0} & \hphantom{2\,\texttimes\,}239.2\hphantom{\,+\,000.0} & 1.35 \textcolor{Green}{(\textminus0.03)} & 313.6 \textcolor{Green}{(+7.4)} \\
\cmidrule{2-8}
& \multicolumn{2}{@{}l}{SiT-XL/2-REPA-E} & E2E-VA-VAE & 674.9\,+\,41.4 & \textcolor{gray!70}{2\,\texttimes}\,237.3\,+\,252.2 & 1.12 \hphantom{(+0.00)} & 302.9 \hphantom{(+0.0)} \\
\rowcolor{blue!6} \cellcolor{white} & & \textbf{DWD} & -- & 678.5\hphantom{\,+\,00.0} & \hphantom{2\,\texttimes\,}237.7\hphantom{\,+\,000.0} & 1.06 \textcolor{Green}{(\textminus0.06)} & 303.1 \textcolor{Green}{(+0.2)} \\
\multirow{-6}{*}[\dimexpr(\aboverulesep+\belowrulesep+\cmidrulewidth)/2\relax]{\rotatebox[origin=c]{90}{\textbf{Latent models}}} & & \emph{DWD-conv} & -- & 679.0\hphantom{\,+\,00.0} & \hphantom{2\,\texttimes\,}239.2\hphantom{\,+\,000.0} & 1.10 \textcolor{Green}{(\textminus0.02)} & 305.7 \textcolor{Green}{(+2.8)} \\
\midrule
& \multicolumn{2}{@{}l}{JiT-H/16} & -- & 952.5\hphantom{\,+\,00.0} & \textcolor{gray!70}{2\,\texttimes}\,363.9\hphantom{\,+\,000.0} & 1.90 \hphantom{(+0.00)} & 300.9 \hphantom{(+0.0)} \\
\rowcolor{blue!6} \cellcolor{white} & & \textbf{DWD} & -- & 956.7\hphantom{\,+\,00.0} & \hphantom{2\,\texttimes\,}363.9\hphantom{\,+\,000.0} & 1.85 \textcolor{Green}{(\textminus0.05)} & 305.7 \textcolor{Green}{(+4.8)} \\
& & \emph{DWD-conv} & -- & 957.2\hphantom{\,+\,00.0} & \hphantom{2\,\texttimes\,}365.4\hphantom{\,+\,000.0} & 1.87 \textcolor{Green}{(\textminus0.03)} & 305.3 \textcolor{Green}{(+4.4)} \\
& & \emph{w/o separate head} & -- & 955.8\hphantom{\,+\,00.0} & \hphantom{2\,\texttimes\,}363.9\hphantom{\,+\,000.0} & 1.91 \textcolor{BrickRed}{(+0.01)} & 302.4 \textcolor{Green}{(+1.5)} \\
\cmidrule{2-8}
& \multicolumn{2}{@{}l}{AnisoFlow-H/16} & -- & 953.1\hphantom{\,+\,00.0} & \textcolor{gray!70}{2\,\texttimes}\,364.8\hphantom{\,+\,000.0} & 1.55 \hphantom{(+0.00)} & 300.1 \hphantom{(+0.0)} \\
\rowcolor{blue!6} \cellcolor{white} & & \textbf{DWD} & -- & 957.3\hphantom{\,+\,00.0} & \hphantom{2\,\texttimes\,}363.9\hphantom{\,+\,000.0} & 1.43 \textcolor{Green}{(\textminus0.12)} & 303.5 \textcolor{Green}{(+3.4)} \\
& & \emph{DWD-conv} & -- & 957.8\hphantom{\,+\,00.0} & \hphantom{2\,\texttimes\,}365.4\hphantom{\,+\,000.0} & 1.44 \textcolor{Green}{(\textminus0.11)} & 303.6 \textcolor{Green}{(+3.5)} \\
\multirow{-8}{*}[\dimexpr(\aboverulesep+\belowrulesep+\cmidrulewidth)/2\relax]{\rotatebox[origin=c]{90}{\textbf{Pixel models}}} & & \emph{w/o separate head} & -- & 956.4\hphantom{\,+\,00.0} & \hphantom{2\,\texttimes\,}364.8\hphantom{\,+\,000.0} & 1.71 \textcolor{BrickRed}{(+0.16)} & 302.9 \textcolor{Green}{(+2.8)} \\
\bottomrule
\end{tabular}
}
\end{table}

\textbf{Improvement over latent diffusion baselines.}
Removing the decoder improves both latent baselines (Table~\ref{tab:latent}): FID improves from 1.38 to 1.36 for REPA and from 1.12 to 1.06 for REPA-E, with IS improving as well. The two baselines use different latent spaces and decoders, yet a linear head improves on both. Removing the decoder also saves 622 decoder GFLOPs for REPA, 2.6\texttimes\ the cost of a single denoiser evaluation, and 252 decoder GFLOPs for REPA-E.

\textbf{Improvement over pixel diffusion baselines.}
The pixel models have no decoder to remove, so DWD only adds a second output trained with pixel quality losses. As shown in Table~\ref{tab:latent}, DWD improves JiT from 1.90 to 1.85 FID and AnisoFlow from 1.55 to 1.43, a new state of the art among pixel flow models, while raising IS by 3--5 points. This also narrows the FID gap between pixel and latent models on ImageNet.

\textbf{Separate head vs.\ shared head.}
The pixel-model gains have two sources: the pixel quality losses, which latent decoders are trained with but pixel models normally lack, and the separate head that receives them. The \emph{w/o separate head} rows in Table~\ref{tab:latent} isolate the latter by fusing the two heads in both training and inference, so the denoising head receives both pixel losses and trajectory anchoring at their respective timesteps, with everything else unchanged. This degrades baseline FID (1.55 to 1.71 for AnisoFlow), so the losses alone do not account for the gain. The degradation reflects a conflict: the $\ell_2$ denoising loss favors mode coverage, whereas LPIPS and adversarial losses favor perceptual quality but collapse diversity (better IS but worse FID than the baseline). Trajectory anchoring cannot fully counteract this interference, so the pixel losses need a separate head.

\textbf{Linear vs. convolutional head.}
We ablate the capacity of the linear DWD head by adding a small convolutional residual refiner conditioned on the linear prediction and the last-layer tokens (\emph{DWD-conv}; Appendix~\ref{app:implementation}). This extension brings no consistent improvement across the ImageNet models in Table~\ref{tab:latent}. A linear head therefore suffices to obtain the quality gains under our tested settings.

\begin{figure}[t]
\hfill
\centering
\begin{minipage}[t]{0.32\linewidth}
\centering
\small \textbf{(a)} \textbf{ImageNet FID vs.\ $t_{\mathrm{final}}$}\\[3pt]
\includegraphics[width=0.9\linewidth]{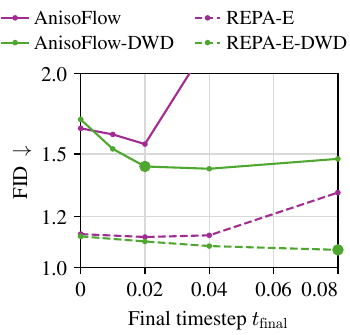}
\end{minipage}\hspace{0.04\linewidth}
\hfill
\begin{minipage}[t]{0.58\linewidth}
\centering
\small \textbf{(b)} \textbf{Same-seed crops (AnisoFlow-L/16-T2I pixel diffusion)}\\[3pt]
\includegraphics[width=0.9\linewidth]{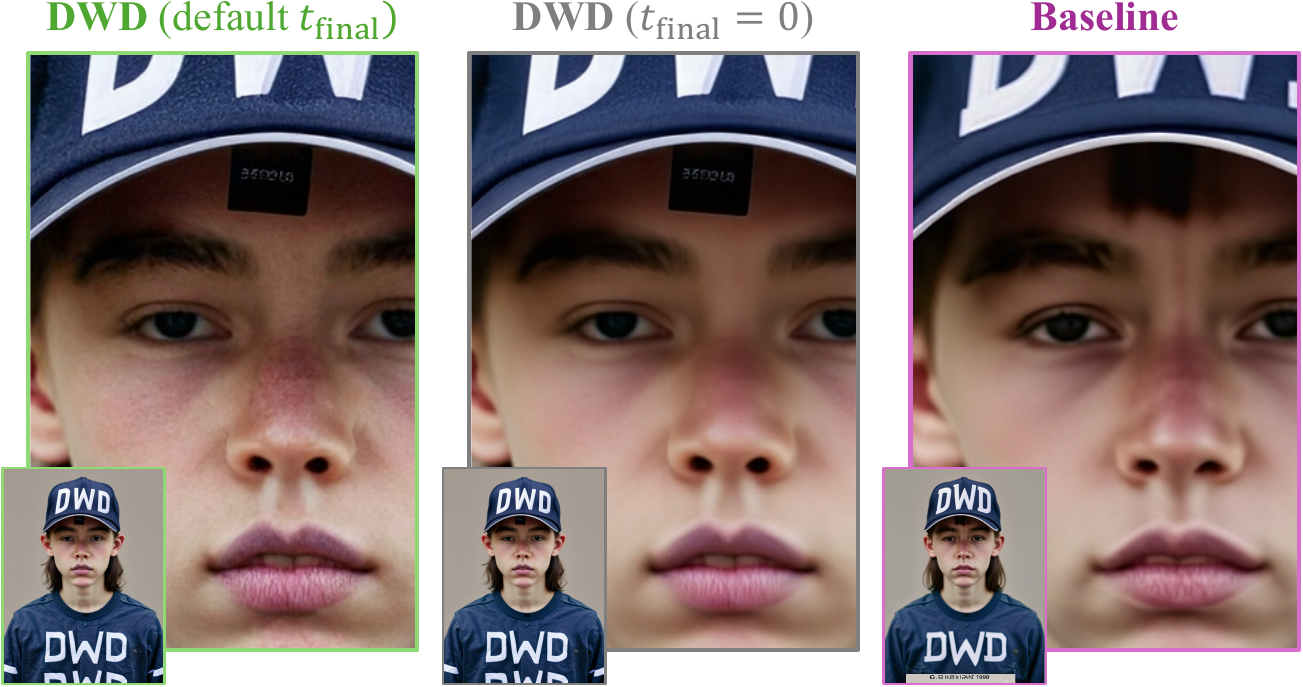}
\end{minipage}
\caption{\textbf{A nonzero final timestep is necessary for the DWD quality gain.} \textbf{(a)} At $t_{\mathrm{final}}{=}0$, where the DWD head receives a fully denoised input, FID matches or trails the baseline's, so the gain vanishes; a small positive timestep recovers it. Larger circles mark the defaults used in Table~\ref{tab:latent}. \textbf{(b)} Qualitative results also confirm the trend: DWD at its default timestep recovers skin and fabric texture, while DWD at $t_{\mathrm{final}}{=}0$ is as smooth as the baseline. The large panels are zoomed crops, with the full image inset.}
\label{fig:terminal}
\hfill
\end{figure}

\textbf{Nonzero final timestep.}
\label{sec:exp_terminal}
Fig.~\ref{fig:terminal} sweeps the final timestep at which the DWD head is evaluated. Quality degrades toward $t{=}0$, where the input is the fully denoised state a decoder would receive, and recovers at a small positive timestep. This shows that the DWD head needs a partially denoised input to produce high-quality pixels, rather than mimicking a decoder applied to a clean state.

\subsection{Text-to-image generation}
\label{sec:exp_t2i}

We show that DWD replaces the decoder with matched quality on latent baselines, including a few-step distilled model, and recovers photorealistic texture detail on the pixel baseline. We also ablate the adversarial loss to show that it is not critical on text-to-image generation.

\textbf{Setup.}
We test two latent models built on FLUX.2 klein 9B~\citep{flux2}, the 32-step Base model and its distilled 4-step variant, and one pixel model, AnisoFlow-L/16-T2I~\citep{\anisoflowbibkey} (0.9B). All DWD extensions are finetuned on public real-image datasets~\citep{textatlas,redcaps,laion} at mixed aspect ratios. We evaluate at 1024\texttimes1024 on COCO-10k~\citep{coco,sdxllightning} (FID, patch FID (pFID)~\citep{sdxllightning}, HPSv3~\citep{hpsv3}, HPSv2.1~\citep{hpsv2}), OneIG-Bench~\citep{oneig}, and LongText-Bench~\citep{qwen}. Training and sampling details are in Appendix~\ref{app:implementation}.

\begin{table}[t]
\vspace{-2ex}
\centering
\caption{\textbf{Results of text-to-image generation (1024\texttimes1024).} {\setlength{\fboxsep}{1pt}\colorbox{blue!6}{Shaded rows}} are our DWD extensions, finetuned from the baseline above them.}
\label{tab:t2i}
\resizebox{0.9\linewidth}{!}{%
\setlength{\tabcolsep}{0.3em}
\begin{tabular}{@{}>{\columncolor{white}[0pt][0pt]}c@{\hspace{8pt}}lcccccccc@{}}
\toprule
& \multirow{2}[5]{*}{\textbf{Method}} & \multirow{2}[5]{*}{\makecell[c]{\textbf{Params (B)} \\ (+\,decoder)}} & \multirow{2}[5]{*}{\makecell[c]{\textbf{TFLOPs} \\ (final step)}} & \multicolumn{4}{c}{COCO-10k} & \multirow{2}[5]{*}{\textbf{OneIG}\textuparrow} & \multirow{2}[5]{*}{\textbf{LongText}\textuparrow} \\
\cmidrule(lr){5-8}
& & & & \footnotesize\textbf{FID}\textdownarrow & \footnotesize\textbf{pFID}\textdownarrow & \footnotesize\textbf{HPSv3}\textuparrow & \footnotesize\makecell[c]{\textbf{HPS} \\ \textbf{v2.1}\textuparrow} & & \\
\midrule
& FLUX.2 klein Base 9B & 9.08\,+\,0.05 & \textcolor{gray!70}{2\,\texttimes}\,75.51\,+\,10.47 & 15.13 & 18.69 & 10.54 & \textbf{0.284} & \textbf{0.556} & \textbf{0.909} \\
\rowcolor{blue!6} \cellcolor{white}  & \multicolumn{1}{@{}>{\columncolor{blue!6}[0pt][\tabcolsep]}l}{\quad\textbf{DWD}} & 9.15\hphantom{\,+\,0.00} & \hphantom{2\,\texttimes\,}75.54\hphantom{\,+\,00.00} & \textbf{15.05} & \textbf{18.38} & 10.54 & \textbf{0.284} & \textbf{0.556} & \multicolumn{1}{>{\columncolor{blue!6}[\tabcolsep][0pt]}c@{}}{0.908} \\
& \quad\emph{w/o adversarial loss} & 9.15\hphantom{\,+\,0.00} & \hphantom{2\,\texttimes\,}75.54\hphantom{\,+\,00.00} & 15.10 & 18.97 & \textbf{10.57} & \textbf{0.284} & 0.555 & 0.907 \\
\cmidrule{2-10}
& FLUX.2 klein 9B (4-step) & 9.08\,+\,0.05 & \hphantom{2\,\texttimes\,}75.51\,+\,10.47 & \textbf{21.50} & 25.52 & \textbf{12.23} & \textbf{0.307} & 0.535 & 0.867 \\
\rowcolor{blue!6} \cellcolor{white} \multirow{-5}{*}[\dimexpr(\aboverulesep+\belowrulesep+\cmidrulewidth)/2\relax]{\rotatebox[origin=c]{90}{\textbf{Latent}}} & \multicolumn{1}{@{}>{\columncolor{blue!6}[0pt][\tabcolsep]}l}{\quad\textbf{DWD}} & 9.15\hphantom{\,+\,0.00} & \hphantom{2\,\texttimes\,}75.54\hphantom{\,+\,00.00} & 21.92 & \textbf{25.33} & 12.15 & \textbf{0.307} & \textbf{0.537} & \multicolumn{1}{>{\columncolor{blue!6}[\tabcolsep][0pt]}c@{}}{\textbf{0.871}} \\
\midrule
& AnisoFlow-L/16-T2I & 0.92\hphantom{\,+\,0.00} & \textcolor{gray!70}{2\,\texttimes}\,5.99\hphantom{\,+\,0.00} & 28.86 & 36.59 & \textbf{10.72} & 0.297 & 0.528 & 0.850 \\
\rowcolor{blue!6} \cellcolor{white}  & \multicolumn{1}{@{}>{\columncolor{blue!6}[0pt][\tabcolsep]}l}{\quad\textbf{DWD}} & 0.94\hphantom{\,+\,0.00} & \hphantom{2\,\texttimes\,}5.98\hphantom{\,+\,0.00} & \textbf{27.54} & \textbf{32.26} & 10.63 & \textbf{0.301} & \textbf{0.533} & \multicolumn{1}{>{\columncolor{blue!6}[\tabcolsep][0pt]}c@{}}{\textbf{0.852}} \\
\multirow{-3}{*}{\rotatebox[origin=c]{90}{\textbf{Pixel}}} & \quad\emph{w/o adversarial loss} & 0.94\hphantom{\,+\,0.00} & \hphantom{2\,\texttimes\,}5.98\hphantom{\,+\,0.00} & 28.93 & 34.52 & 10.67 & 0.299 & 0.532 & 0.843 \\
\bottomrule
\end{tabular}
}
\end{table}

\begin{figure}[t]
\centering
\includegraphics[width=0.9\linewidth]{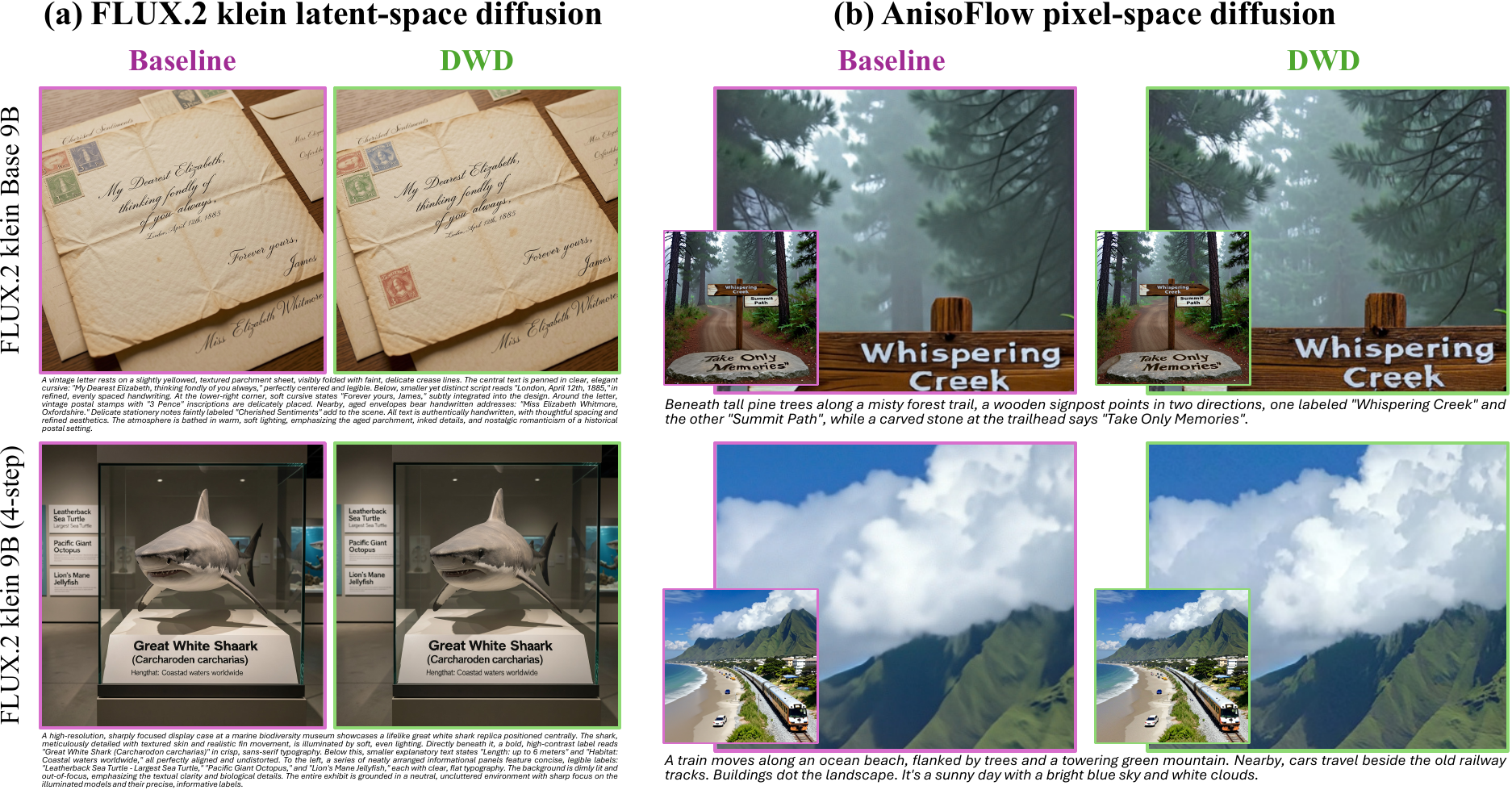}
\caption{\textbf{Text-to-image comparisons.} Each pair shows the baseline (left) and its DWD extension (right) for the same prompt and seed. \textbf{(a)} On the latent FLUX.2 klein models, DWD closely matches the baseline's content and texture without the decoder (zoom in to view detail). \textbf{(b)} On the pixel AnisoFlow model, DWD preserves the composition while recovering photorealistic texture in bark, foliage, and clouds. The large panels are zoomed crops, with the full image inset.}
\label{fig:t2i_qual}
\end{figure}

\textbf{Decoder removal without quality loss.}
On FLUX.2 klein Base, DWD and the baseline are effectively tied on every metric in Table~\ref{tab:t2i}, and same-seed samples are nearly indistinguishable (Fig.~\ref{fig:t2i_qual}(a)). This is expected: the FLUX.2 latent is much wider than those of other VAEs, with 128 latent dimensions per token for a 768-dimensional pixel patch, compared with 64 for a 3072-dimensional patch in Wan, so decoding from it is easy, leaving little room for improvement. The result nevertheless shows that a linear head can replace it, removing 10.47 TFLOPs of decoder computation per image.

\textbf{Compatibility with few-step model.}
The distilled 4-step FLUX.2 klein tests the few-step regime, where the decoder takes a larger share of the total inference cost. With only the timestep ranges adjusted (Appendix~\ref{app:finetuning}), DWD is again effectively tied with the baseline in Table~\ref{tab:t2i} and visually indistinguishable in Fig.~\ref{fig:t2i_qual}(a), even though its final step operates at a much larger timestep ($t{=}0.7$ after shifting) than in the 32-step model.

\textbf{Improved detail over pixel diffusion baseline.}
Pixel diffusion models tend to fall short of latent models in very fine texture, even with very large denoisers~\citep{hidreamo1} or AnisoFlow's improved parameterization~\citep{\anisoflowbibkey}. With a linear head and 3k finetuning iterations, DWD recovers photorealistic skin, hair, foliage, and cloud detail while preserving composition (Fig.~\ref{fig:teaser}(b), Fig.~\ref{fig:terminal}(b), Fig.~\ref{fig:t2i_qual}(b)). The gain also shows up in the low-level metrics in Table~\ref{tab:t2i}: COCO FID improves from 28.86 to 27.54 and pFID from 36.59 to 32.26, the latter showing a more significant improvement due to its sensitivity to fine detail. The remaining higher-level metrics are effectively tied, as the trajectory is preserved.

\textbf{Adversarial loss.}
Without the adversarial loss (Table~\ref{tab:t2i}), FLUX.2 klein and AnisoFlow's DWD extensions stay close to full DWD on most metrics, except that pFID worsens more noticeably, though still close to or outperforming the baseline. The adversarial loss is thus helpful but not critical, in line with recent decoder research that drops it~\citep{qwen,qwenvae2}.

\subsection{Text-to-video generation}
\label{sec:exp_video}

We apply DWD to the Wan2.1 14B video model, where each token must predict 4 frames rather than a single image patch. Despite this harder setting, DWD not only removes the decoder but visibly reduces distortion and flickering, lowering warping error by at least 23\%.

\textbf{Setup.}
We finetune Wan2.1 T2V 14B~\citep{wan} on Vchitect T2V DataVerse~\citep{vchitect} clips at mixed 360p, 480p, and 720p resolutions, all downsampled from sources of at least 720p to obtain sharper real pixels (Appendix~\ref{app:finetuning}). We evaluate 81-frame 832\texttimes480 generations on extended VBench prompts~\citep{vbench} from \citet{selfforcing} and report warping error~\citep{lai2018consistency}, frame-level HPSv3 and HPSv2.1, and seven VBench quality dimensions. For both the baseline and DWD, we additionally evaluate a quantized model with FP8 weights and activations (W8A8) and 8-bit SageAttention2 kernels~\citep{zhang2025sageattention2}, which better reflects efficient deployment settings.

\begin{table}[t]
\centering
\caption{\textbf{Results of text-to-video generation (832\texttimes480, 81 frames) on VBench.} {\setlength{\fboxsep}{1pt}\colorbox{blue!6}{Shaded rows}} are our DWD extensions, finetuned from the baseline above them. \emph{Finetuned decoder} keeps the baseline DiT and finetunes its decoder on the same data with the same pixel losses as DWD (Appendix~\ref{app:decoder_baseline}). The 8-bit baseline and its DWD extension both use FP8 W8A8 quantization with 8-bit SageAttention2~\citep{zhang2025sageattention2}.}
\label{tab:video}
\resizebox{\linewidth}{!}{%
\setlength{\tabcolsep}{0.2em}
\begin{tabular}{@{}lcccccccccccc@{}}
\toprule
\multirow{2}[5]{*}{\textbf{Method}} & \multirow{2}[5]{*}{\makecell[c]{\textbf{Params (B)} \\ (+\,decoder)}} & \multirow{2}[5]{*}{\makecell[c]{\textbf{PFLOPs} \\ (final step)}} & \multirow{2}[5]{*}{\makecell[c]{\textbf{Warping} \\ \textbf{error} \\ (\texttimes10\textsuperscript{-3})\textdownarrow}} & \multicolumn{2}{c}{Image metrics} & \multicolumn{7}{c}{VBench dimensions} \\
\cmidrule(lr){5-6} \cmidrule(l){7-13}
& & & & \footnotesize\textbf{HPSv3}\textuparrow & \footnotesize\makecell[c]{\textbf{HPS} \\ \textbf{v2.1}\textuparrow} & \footnotesize\makecell[c]{\textbf{Subject} \\ \textbf{consist.}\textuparrow} & \footnotesize\makecell[c]{\textbf{Backgr.} \\ \textbf{consist.}\textuparrow} & \footnotesize\makecell[c]{\textbf{Temporal} \\ \textbf{flicker.}\textuparrow} & \footnotesize\makecell[c]{\textbf{Motion} \\ \textbf{smooth.}\textuparrow} & \footnotesize\makecell[c]{\textbf{Dynamic} \\ \textbf{degree}\textuparrow} & \footnotesize\makecell[c]{\textbf{Aesthetic} \\ \textbf{quality}\textuparrow} & \footnotesize\makecell[c]{\textbf{Imaging} \\ \textbf{quality}\textuparrow} \\
\midrule
Wan2.1 14B & 14.29\,+\,0.07 & \textcolor{gray!70}{2\,\texttimes}\,1.68\,+\,0.28 & 1.52 & \textbf{9.06} & \textbf{0.266} & 0.951 & 0.981 & 0.990 & 0.983 & \textbf{0.736} & 0.682 & \textbf{0.677} \\
\rowcolor{blue!6} \multicolumn{1}{@{}>{\columncolor{blue!6}[0pt][\tabcolsep]}l}{\quad\textbf{DWD}} & 14.49\hphantom{\,+\,0.00} & \hphantom{2\,\texttimes\,}1.68\hphantom{\,+\,0.00} & \textbf{1.18} & 9.04 & \textbf{0.266} & \textbf{0.952} & \textbf{0.982} & \textbf{0.993} & \textbf{0.984} & 0.711 & \textbf{0.685} & \multicolumn{1}{>{\columncolor{blue!6}[\tabcolsep][0pt]}c@{}}{0.671} \\
\quad\emph{finetuned decoder} & 14.29\,+\,0.07 & \textcolor{gray!70}{2\,\texttimes}\,1.68\,+\,0.28 & 1.39 & 9.03 & \textbf{0.266} & 0.951 & \textbf{0.982} & 0.992 & 0.983 & 0.731 & 0.681 & 0.676 \\
\midrule
Wan2.1 14B (8-bit) & 14.29\,+\,0.07 & \textcolor{gray!70}{2\,\texttimes}\,1.68\,+\,0.28 & 1.69 & 8.99 & 0.265 & 0.950 & 0.979 & 0.990 & 0.982 & \textbf{0.728} & 0.675 & \textbf{0.676} \\
\rowcolor{blue!6} \multicolumn{1}{@{}>{\columncolor{blue!6}[0pt][\tabcolsep]}l}{\quad\textbf{DWD}} & 14.49\hphantom{\,+\,0.00} & \hphantom{2\,\texttimes\,}1.68\hphantom{\,+\,0.00} & \textbf{1.20} & \textbf{9.01} & \textbf{0.267} & \textbf{0.951} & \textbf{0.982} & \textbf{0.993} & \textbf{0.984} & 0.703 & \textbf{0.683} & \multicolumn{1}{>{\columncolor{blue!6}[\tabcolsep][0pt]}c@{}}{0.668} \\
\bottomrule
\end{tabular}
}
\end{table}

\begin{figure}[t]
\centering
\includegraphics[width=0.98\linewidth]{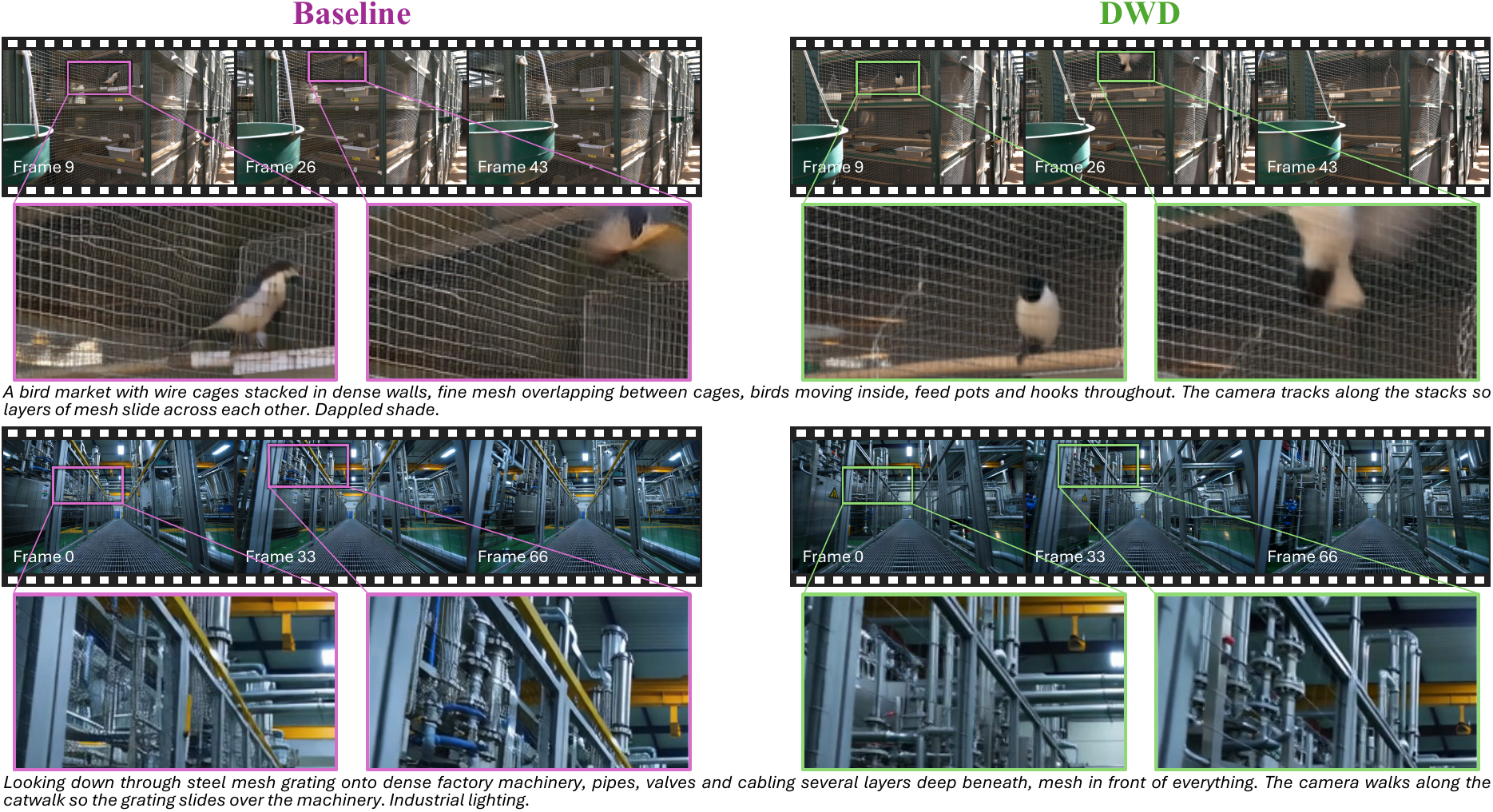}
\caption{\textbf{Text-to-video comparisons (BF16).} Three frames from the baseline Wan2.1 14B (left) and its DWD extension (right) for the same prompt and seed, with two zoomed crops per strip. The baseline decoder warps the wire mesh into wavy, uneven lines (top) and breaks the dense pipes into speckle (bottom); neither defect is static, both shifting from frame to frame. DWD keeps the mesh regular and the pipe structure intact, and holds them steady as the camera tracks across the scene.}
\label{fig:video_qual}
\end{figure}

\textbf{Decoder removal with reduced distortion and flickering.}
DWD removes the 0.28-PFLOP Wan decoder and replaces it with a linear head at negligible cost (Table~\ref{tab:video}), while also improving visual quality. The baseline decoder distorts fine detail such as small text and dense structures, and the distortion changes from frame to frame, which appears as flicker (Fig.~\ref{fig:teaser}(a) and Fig.~\ref{fig:video_qual}). With the decoder removed, the linear DWD head keeps these details stable and legible, noticeably reducing flickering\ifarxiv{} (see the project page for video comparisons).\else{} (see the supplementary material for video comparisons).\fi{} Warping error captures this artifact directly: it aligns consecutive frames by optical flow, masks occlusions, and measures the remaining photometric error, which DWD reduces by 23\% (1.52 to 1.18) in Table~\ref{tab:video}. VBench's temporal flickering score is less suitable here, since it is the raw pixel difference between consecutive frames without accounting for residual motion, and it saturates near its maximum for both models. The remaining VBench dimensions and HPS metrics are effectively tied, except for a slightly lower dynamic degree, though qualitative results show no visible motion reduction. A decoder finetuned on the same data with the same pixel losses reduces warping error only to 1.39 (Table~\ref{tab:video} and Appendix~\ref{app:decoder_baseline}), so most of the gain comes from DWD rather than from the training data.

\textbf{8-bit inference.}
Quantization degrades the baseline noticeably (warping error 1.52 to 1.69) but barely affects DWD (1.18 to 1.20), widening the warping error gap to 29\% in Table~\ref{tab:video}. The other metrics follow the same pattern as under BF16. The DWD extension is thus robust to reduced-precision inference, and the quality gain over the decoder is larger in the deployment setting.

\section{Conclusion}
\label{sec:conclusion}

DWD replaces the decoder with a linear pixel head on the diffusion transformer, finetuning only the last few blocks and output heads. Across latent image and video generators, DWD preserves or improves pixel quality while removing the decoder entirely. On video, this visibly reduces detail distortion and flickering. On pixel diffusion models, a separate DWD head recovers photorealistic texture that the denoising head oversmooths. The broader lesson is that final pixel prediction should be separated from intermediate denoising, so that pixel-space objectives can improve quality without interfering with the sampling trajectory.

\textbf{Limitations.}
The DWD quality gain shrinks when the decoder is already strong, as on FLUX.2.
We have not proved that the linear head is sufficient for spatial patch sizes beyond 16\texttimes16 pixels, where a stronger head like DWD-conv may be needed. Also, we have not tuned the pixel losses per model, leaving potential room for improvement.

\section*{AI Use Statement}
In this work, we used generative AI tools (LLM-based assistants) to edit the paper for grammar and readability and to accelerate the implementation of our experiments\ifarxiv\else{} and the supplementary code\fi. All AI-assisted work was reviewed and verified by the authors, and we take responsibility for the final content of this paper.

\section*{Reproducibility Statement}
\ifarxiv\else We provide a minimal, self-contained implementation of DWD in the supplementary material. The full code and finetuned models will be released upon publication. In addition, \fi Appendix~\ref{app:implementation} gives the implementation details: the head and its initialization (Appendix~\ref{app:head}), the per-model finetuning settings and training data (Appendix~\ref{app:finetuning} and Table~\ref{tab:app_settings}), the inference procedure (Appendix~\ref{app:sampling}), the video evaluation protocol (Appendix~\ref{app:evaluation}), and the parameter and FLOP accounting (Appendix~\ref{app:cost}).

\bibliographystyle{iclr2027_conference}
\bibliography{iclr2027_conference}

\clearpage
\appendix
\section{Implementation Details}
\label{app:implementation}

This section complements Sec.~\ref{sec:method} with the specifics of the DWD head, finetuning, inference, and evaluation, with per-model settings collected in Table~\ref{tab:app_settings}.

\subsection{DWD Head}
\label{app:head}

\textbf{Output format.}
The linear head maps each last-layer token to one pixel patch, and the patches are assembled into the image. Image models use 16\texttimes16 patches. Wan uses 4\texttimes16\texttimes16 spatiotemporal patches and a separate 1\texttimes16\texttimes16 head for the first latent frame, which covers a single pixel frame in the Wan VAE. Following AsymFlow~\citep{chen2026asymflow}, latent models and AnisoFlow-L/16-T2I predict pixels in normalized Oklab color space~\citep{oklab}, while the pixel quality losses are applied to the converted RGB values, matching the common autoencoder training recipe. The ImageNet pixel models predict RGB directly, matching their native output.

\textbf{Least-squares initialization.}
The head is initialized by linear regression from last-layer tokens to ground-truth pixel patches. We run the frozen pretrained model on 10k training images with timesteps drawn from $p_{\mathrm{pix}}$, the same distribution used for the pixel losses, and solve the normal equations for the weight and bias jointly. The same procedure produces the probes in Fig.~\ref{fig:probe}, where the latent probe regresses from the denoised latent instead of the tokens.

\textbf{Convolutional head (DWD-conv).}
The DWD-conv ablation keeps the linear head and adds a residual correction, following lightweight super-resolution networks~\citep{shi2016espcn,yu2018wdsr,ayazoglu2021xlsr}: the nonlinear convolutions run at half resolution, a final sub-pixel convolution (pixel shuffle) restores the full resolution, and the branch predicts a residual over a linear base prediction. It packs each 2\texttimes2 neighborhood of the linear pixel prediction into channels and refines this packed grid with a convolutional branch. The last-layer tokens are projected through a 128-dimensional bottleneck and expanded into a 32-channel spatial feature map, which is added to a 1\texttimes1 projection of the packed pixels. Two residual blocks with 3\texttimes3 convolutions and SiLU activations predict the correction through a zero-initialized output convolution, so the head starts as the linear head. For video, the convolutions operate on each frame independently.

\subsection{Finetuning}
\label{app:finetuning}

\begin{table}[h]
\centering
\caption{\textbf{DWD finetuning settings per model.} $K$ is the number of finetuned blocks out of the total. Adv.\ start is the iteration at which the generator adversarial loss is switched on, after a discriminator warmup. $\lambda$ is the trajectory anchoring weight in Eq.~(\ref{eq:total_loss}), and $p_{\mathrm{traj}}$ and $p_{\mathrm{pix}}$ are the timestep ranges of the two losses before the timestep shift. Mixed-resolution runs use the batch sizes given in Appendix~\ref{app:finetuning}, and FLUX.2 klein Base uses the model's resolution-dependent shift (dyn.). GPU hours are estimated for H100 GPUs.}
\label{tab:app_settings}
\resizebox{\linewidth}{!}{%
\setlength{\tabcolsep}{0.45em}
\begin{tabular}{lccccccccccc}
\toprule
\textbf{Model} & $K$ & \textbf{Iters} & \textbf{Batch} & \textbf{Adv.\ start} & $\lambda$ & $p_{\mathrm{traj}}$ & $p_{\mathrm{pix}}$ & \textbf{Shift} & \textbf{Final $t$} & \textbf{Pixel space} & \textbf{GPU hours} \\
\midrule
SiT-XL/2-REPA & 16 / 28 & 100k & 1024 & 52.5k & 5 & $[0.04, 1]$ & $[0, 0.1]$ & 1 & 0.08 & Oklab & 170 \\
SiT-XL/2-REPA-E & 8 / 28 & 200k & 1024 & 152.5k & 10 & $[0.04, 1]$ & $[0, 0.1]$ & 1 & 0.08 & Oklab & 200 \\
JiT-H/16 & 2 / 32 & 4k & 1024 & 2.5k & 50 & $[0.04, 1]$ & $[0, 0.1]$ & 1 & sampler & RGB & 8 \\
AnisoFlow-H/16 & 2 / 32 & 4k & 1024 & 2.5k & 50 & $[0.04, 1]$ & $[0, 0.1]$ & 1 & sampler & RGB & 8 \\
\midrule
FLUX.2 klein Base 9B & 4 / 32 & 26k & mixed & 21k & 10 & $[0.04, 1]$ & $[0, 0.1]$ & dyn. & 0.3 & Oklab & 350 \\
FLUX.2 klein 9B (4-step) & 4 / 32 & 26k & 128 & 21k & 10 & $[0.3, 1]$ & $[0, 0.5]$ & 7 & sampler & Oklab & 340 \\
AnisoFlow-L/16-T2I & 4 / 23 & 3k & 128 & 2k & 500 & $[0.04, 1]$ & $[0, 0.1]$ & 5 & sampler & Oklab & 14 \\
\midrule
Wan2.1 T2V 14B & 8 / 40 & 40k & mixed & 21k & 5 & $[0.04, 1]$ & $[0.02, 0.1]$ & 7 & 0.3 & Oklab & 4300 \\
Wan2.1 T2V 1.3B (4-step) & 8 / 30 & 40k & mixed & 35k & 5 & $[0.3, 0.94]$ & $[0.02, 0.3]$ & 5.5 & sampler & Oklab & 980 \\
\bottomrule
\end{tabular}
}
\end{table}

\textbf{Trainable components.}
Table~\ref{tab:app_settings} lists the number of finetuned blocks. The selected blocks are trained in full, except for the SiT models, where we freeze the query and key projections because SiT has no QK normalization and finetuning them can be unstable. All models also train both output heads, the final modulation, and the duplicated timestep embedding MLP (together with the class, guidance, or text modulation MLPs where present), all initialized from the pretrained weights.

\textbf{Pixel losses.}
The $\ell_1$ and LPIPS terms are weighted $1{:}1$ in all runs. The discriminator follows the StyleGAN-T design~\citep{sauer2023stylegant} as adopted by RAE~\citep{zheng2025rae}. For the text-to-image and video models, it scores multiple random 256\texttimes256 crops of each image or frame rather than the full resolution. For video, the discriminator features are extracted per frame, and we insert a residual block with a 1D convolution over the frame axis (kernel size 3) before the per-token scoring. The adversarial weight is set adaptively to match the gradient norm of the $\ell_1$ and LPIPS loss at the predicted pixels, following the gradient balancing of \citet{esser2021vqgan} but measured at the pixels rather than at the last layer's weights. The discriminator is trained for a short warmup before the generator adversarial loss is switched on at the iteration listed in Table~\ref{tab:app_settings} (2.5k iterations of warmup for the ImageNet models, 1k for the others).

\textbf{Optimization.}
All runs use Adam~\citep{adam} with learning rate $10^{-4}$, betas $(0.9, 0.95)$, no weight decay, a short linear warmup, and a 5\texttimes\ learning rate for the DWD head. Weights are evaluated with the exponential moving average of \citet{karras2024analyzingimprovingtrainingdynamics} ($\gamma=7$), and Wan additionally caps the EMA momentum at $0.998$. For FLUX.2 klein and Wan, the frozen blocks are quantized to FP8 during finetuning to reduce memory and speed up training. The frozen reference model is also quantized to FP8, while the trainable layers are trained in FP32 with BF16 mixed precision. The trainable layers therefore learn from a slightly degraded FP8 reference, which works against DWD rather than favoring its 8-bit results. All main and qualitative results use BF16 inference.

\textbf{Training data.}
The ImageNet models are finetuned on the ImageNet training set at 256\texttimes256. The text-to-image models are finetuned on the one-megapixel training data mix of AnisoFlow~\citep{\anisoflowbibkey}, consisting of TextAtlas~\citep{textatlas}, RedCaps~\citep{redcaps}, and LAION-Aesthetics~\citep{laion} images, with captions from i1~\citep{zeng2026i1} and AsymFlow~\citep{chen2026asymflow}. FLUX.2 klein Base additionally trains on the same images downsampled to roughly 0.5 and 0.25 megapixels. The three tiers use per-iteration batch sizes of 128, 256, and 384 images and are sampled in the ratio $1{:}2{:}3$. The video model is finetuned on clips from Vchitect T2V DataVerse~\citep{vchitect} with the dataset's captions. Clips are bucketed by resolution and length: 352\texttimes640 with 81 or 33 frames (per-iteration batch sizes of 64 and 192 clips), 480\texttimes832 with 81, 49, or 25 frames (32, 64, and 128 clips), and 720\texttimes1280 with 33 or 17 frames (32 and 64 clips). Only source videos of at least 720p are used, so every training clip is downsampled from a higher or equal source resolution rather than upscaled from low-bitrate video, which we found essential for a sharp ground truth.

\subsection{Inference}
\label{app:sampling}

\textbf{Guidance and final step.}
Sampling follows each baseline's default sampler and guidance, with one exception: the SiT-XL/2-REPA DWD extension uses guidance scale 2.3 on the interval $[0, 0.56]$. The final DWD step is evaluated once with the conditional input only, without a classifier-free guidance pair. Since latent models favor a larger final timestep (Fig.~\ref{fig:terminal}(a)), their final timestep is set manually as in Table~\ref{tab:app_settings}, whereas the pixel models and the 4-step models use the last timestep of the baseline sampler. For Wan2.1 14B, the final timestep $t=0.3$ after a shift of 7 corresponds to $t\approx0.058$ before shifting. Its pixel loss skips timesteps below 0.02 before shifting, where Wan's own predictions show strong color drift and detail degradation (common sampling grids do not use these timesteps either).

\textbf{8-bit inference.}
The 8-bit setting quantizes all linear layers to FP8 weights and activations (W8A8) and replaces attention with the 8-bit SageAttention2 kernels~\citep{zhang2025sageattention2}, for both the baseline and DWD.

\subsection{Video Evaluation}
\label{app:evaluation}
Warping error~\citep{lai2018consistency} is computed for every pair of adjacent frames at native resolution: we estimate forward and backward optical flow with RAFT~\citep{teed2020raft}, warp the second frame onto the first, mask occlusions and motion boundaries by forward-backward consistency as in the released scorer of \citet{lai2018consistency}, and average the squared RGB error over valid pixels following \citet{xu2024videogigagan}. Errors are averaged over pairs within each video, then over videos. HPSv3 and HPSv2.1 are computed on five frames sampled uniformly from each video (one frame per second at 16 fps) and averaged.

\subsection{Cost Accounting}
\label{app:cost}
The parameter counts and final-step FLOPs in Tables~\ref{tab:latent}--\ref{tab:video} are computed as follows. Parameter counts include the denoiser and, for baselines, the decoder, excluding text encoders. For DWD extensions, the count includes the duplicated timestep embedding MLP and the linear head, so some DWD models have slightly more parameters than their baselines, though the MLP contributes negligible compute. Final-step FLOPs count one denoiser evaluation, plus the decoder for the baselines, at each model's evaluated resolution. The greyed factor of two marks the baselines' classifier-free guidance pair, which DWD skips in the final step.

\section{Additional Experiments}
\label{app:additional}

\subsection{Decoder-Retaining Baseline}
\label{app:decoder_baseline}
To separate decoder removal from the effect of the training data and pixel losses, we finetune the native Wan2.1 decoder on the same video data with the same $\ell_1$, LPIPS, and adversarial losses as DWD (Appendix~\ref{app:finetuning}), using the same optimizer and EMA settings. The pretrained decoder is already well trained and its training loss stays flat, so we evaluate it after 2k iterations. In Table~\ref{tab:video}, the finetuned decoder reduces warping error only from 1.52 to 1.39, while DWD reduces it to 1.18, and the other metrics are effectively tied. Most of the DWD gain therefore comes from the method rather than from the training data. Finetuning the decoder is also not cheap: one decoder training iteration takes 24\,s, about 2\texttimes\ slower than a DWD iteration on Wan2.1 14B (12.5\,s).

\subsection{Few-Step Video Generation}
\label{app:fewstep_video}
We apply DWD to the 4-step Wan2.1 T2V 1.3B model distilled with rCM~\citep{rcm}, the example from Sec.~\ref{sec:intro}. We finetune its last 8 of 30 blocks on the 352\texttimes640 and 480\texttimes832 tiers of the video data, and keep the sampler's exact timesteps (Table~\ref{tab:app_settings}). The first 32.5k iterations use $p_{\mathrm{traj}}=[0.1, 0.94]$ and $p_{\mathrm{pix}}=[0.02, 0.1]$, after which we switch to the ranges in Table~\ref{tab:app_settings} to match the final timestep of the 4-step sampler. In Table~\ref{tab:app_fewstep}, DWD reduces warping error by 27\% and improves temporal flickering. Note that the 1.3B model is an unfavorable setting for DWD. It is much smaller than Wan2.1 14B, and tuning only 8 blocks leaves limited learnable capacity.

\textbf{Self-distillation.}
With very few steps, the final step has to generate more content, so its pairing with real pixels becomes ambiguous, potentially making $\ell_1$ and perceptual losses less effective. We therefore also test a self-distillation variant, which replaces the real pixels with a target produced by a chain: the reference model's native final step, renoising to a small timestep, and the pixel output of a previously trained small-time DWD model. The results are shown in Table~\ref{tab:app_fewstep}. In this experiment, the small-time DWD model is from the standard run after its first 32.5k iterations. The self-distillation DWD resumes from this checkpoint and continues to 40k iterations. This variant has nearly the same dynamic degree as the baseline (0.822 vs.\ 0.825) and still reduces warping error by 14\%.

\begin{table}[H]
\centering
\caption{\textbf{Few-step video generation on 4-step Wan2.1 T2V 1.3B (BF16).} All models use the same sampler timesteps.}
\label{tab:app_fewstep}
\resizebox{0.98\linewidth}{!}{%
\setlength{\tabcolsep}{0.4em}
\begin{tabular}{@{}lcccccccccc@{}}
\toprule
\multirow{2}[5]{*}{\textbf{Method}} & \multirow{2}[5]{*}{\makecell[c]{\textbf{Warping} \\ \textbf{error} \\ (\texttimes10\textsuperscript{-3})\textdownarrow}} & \multicolumn{2}{c}{Image metrics} & \multicolumn{7}{c}{VBench dimensions} \\
\cmidrule(lr){3-4} \cmidrule(l){5-11}
& & \footnotesize\textbf{HPSv3}\textuparrow & \footnotesize\makecell[c]{\textbf{HPS} \\ \textbf{v2.1}\textuparrow} & \footnotesize\makecell[c]{\textbf{Subject} \\ \textbf{consist.}\textuparrow} & \footnotesize\makecell[c]{\textbf{Backgr.} \\ \textbf{consist.}\textuparrow} & \footnotesize\makecell[c]{\textbf{Temporal} \\ \textbf{flicker.}\textuparrow} & \footnotesize\makecell[c]{\textbf{Motion} \\ \textbf{smooth.}\textuparrow} & \footnotesize\makecell[c]{\textbf{Dynamic} \\ \textbf{degree}\textuparrow} & \footnotesize\makecell[c]{\textbf{Aesthetic} \\ \textbf{quality}\textuparrow} & \footnotesize\makecell[c]{\textbf{Imaging} \\ \textbf{quality}\textuparrow} \\
\midrule
Wan2.1 1.3B (4-step) & 2.84 & 8.32 & \textbf{0.265} & 0.938 & \textbf{0.962} & 0.976 & 0.979 & \textbf{0.825} & \textbf{0.660} & \textbf{0.669} \\
\rowcolor{blue!6} \multicolumn{1}{@{}>{\columncolor{blue!6}[0pt][\tabcolsep]}l}{\quad\textbf{DWD}} & \textbf{2.07} & \textbf{8.46} & 0.258 & \textbf{0.946} & 0.959 & \textbf{0.984} & \textbf{0.983} & 0.811 & 0.657 & \multicolumn{1}{>{\columncolor{blue!6}[\tabcolsep][0pt]}c@{}}{0.648} \\
\rowcolor{blue!6} \multicolumn{1}{@{}>{\columncolor{blue!6}[0pt][\tabcolsep]}l}{\quad\textbf{DWD} (self-distillation)} & 2.45 & 8.45 & 0.261 & 0.942 & 0.956 & 0.982 & 0.981 & 0.822 & 0.656 & \multicolumn{1}{>{\columncolor{blue!6}[\tabcolsep][0pt]}c@{}}{0.631} \\
\bottomrule
\end{tabular}
}
\end{table}

\subsection{End-to-End Latency and Memory}
\label{app:latency}
Table~\ref{tab:app_latency} reports end-to-end latency and peak VRAM for the 4-step models, with FLUX.2 klein 9B at 1024\texttimes1024 and Wan2.1 1.3B at 81 frames of 480\texttimes832, both in eager mode (without \texttt{torch.compile}). DWD makes end-to-end generation 1.12\texttimes\ faster on FLUX.2 and 1.69\texttimes\ faster on Wan, with 5\% and 16\% lower peak VRAM, respectively.

\begin{table}[H]
\centering
\caption{\textbf{End-to-end latency and peak VRAM of 4-step models.}}
\label{tab:app_latency}
\setlength{\tabcolsep}{0.6em}
\scalebox{0.9}{%
\begin{tabular}{@{}lcc@{}}
\toprule
\textbf{Model} & \textbf{Latency (s)}\textdownarrow & \textbf{Peak VRAM (GB)}\textdownarrow \\
\midrule
FLUX.2 klein 9B (4-step) & 0.776 & 37.31 \\
\rowcolor{blue!6} \multicolumn{1}{@{}>{\columncolor{blue!6}[0pt][\tabcolsep]}l}{\quad\textbf{DWD}} & \textbf{0.695} & \multicolumn{1}{>{\columncolor{blue!6}[\tabcolsep][0pt]}c@{}}{\textbf{35.54}} \\
\midrule
Wan2.1 1.3B (4-step) & 4.661 & 19.22 \\
\rowcolor{blue!6} \multicolumn{1}{@{}>{\columncolor{blue!6}[0pt][\tabcolsep]}l}{\quad\textbf{DWD}} & \textbf{2.757} & \multicolumn{1}{>{\columncolor{blue!6}[\tabcolsep][0pt]}c@{}}{\textbf{16.15}} \\
\bottomrule
\end{tabular}
}
\end{table}

\subsection{Comparison with Latent-to-Pixel Adaptation}
\label{app:asymflow}
\begin{figure}[p]
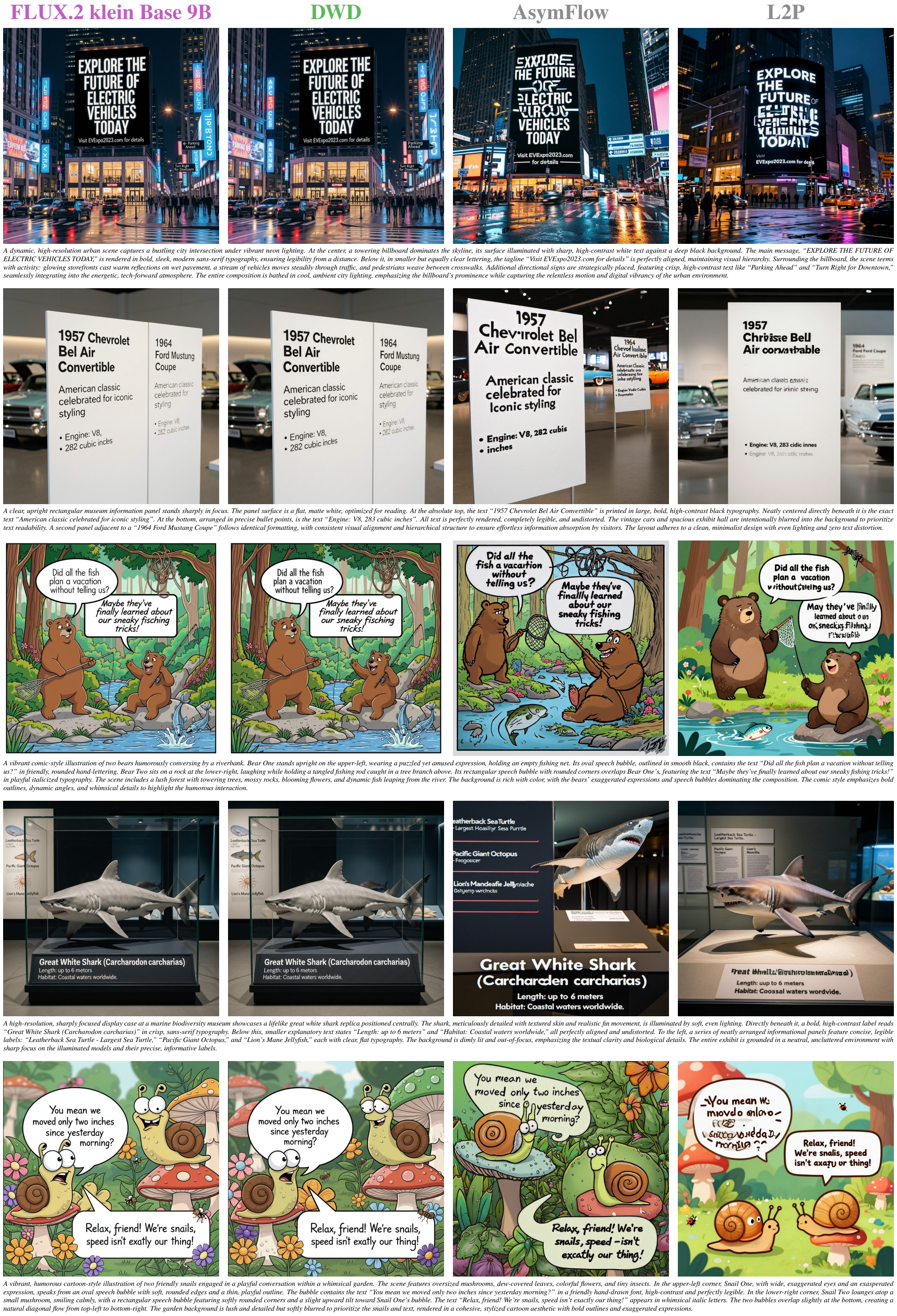

\centering
\begin{minipage}{0.95\linewidth}
\newcommand{\ltimg}[2]{\includegraphics[width=0.242\linewidth]{figures/app_longtext/#1_#2.jpg}}
\newcommand{\lthead}[2]{\makebox[0.242\linewidth][c]{\small\bfseries\textcolor{#1}{#2}}}
\newcommand{\ltrow}[3]{\noindent\ltimg{base}{#1}\hfill\ltimg{dwd}{#1}\hfill\ltimg{asym}{#1}\hfill\ltimg{l2p}{#1}\par\vspace{0.5pt}{\fontsize{3}{3.5}\selectfont\itshape\noindent #2\par}\vspace{#3}}
\noindent\lthead{basecol}{FLUX.2 klein Base 9B}\hfill\lthead{dwdcol}{DWD}\hfill\lthead{detached}{AsymFlow}\hfill\lthead{detached}{L2P}\par\vspace{1pt}
\ltrow{5}{A dynamic, high-resolution urban scene captures a bustling city intersection under vibrant neon lighting. At the center, a towering billboard dominates the skyline, its surface illuminated with sharp, high-contrast white text against a deep black background. The main message, ``EXPLORE THE FUTURE OF ELECTRIC VEHICLES TODAY,'' is rendered in bold, sleek, modern sans-serif typography, ensuring legibility from a distance. Below it, in smaller but equally clear lettering, the tagline ``Visit EVExpo2023.com for details'' is perfectly aligned, maintaining visual hierarchy. Surrounding the billboard, the scene teems with activity: glowing storefronts cast warm reflections on wet pavement, a stream of vehicles moves steadily through traffic, and pedestrians weave between crosswalks. Additional directional signs are strategically placed, featuring crisp, high-contrast text like ``Parking Ahead'' and ``Turn Right for Downtown,'' seamlessly integrating into the energetic, tech-forward atmosphere. The entire composition is bathed in cool, ambient city lighting, emphasizing the billboard's prominence while capturing the relentless motion and digital vibrancy of the urban environment.}{3pt}
\ltrow{116}{A clear, upright rectangular museum information panel stands sharply in focus. The panel surface is a flat, matte white, optimized for reading. At the absolute top, the text ``1957 Chevrolet Bel Air Convertible'' is printed in large, bold, high-contrast black typography. Neatly centered directly beneath it is the exact text ``American classic celebrated for iconic styling''. At the bottom, arranged in precise bullet points, is the text ``Engine: V8, 283 cubic inches''. All text is perfectly rendered, completely legible, and undistorted. The vintage cars and spacious exhibit hall are intentionally blurred into the background to prioritize text readability. A second panel adjacent to a ``1964 Ford Mustang Coupe'' follows identical formatting, with consistent visual alignment and hierarchical structure to ensure effortless information absorption by visitors. The layout adheres to a clean, minimalist design with even lighting and zero text distortion.}{3pt}
\ltrow{592}{A vibrant comic-style illustration of two bears humorously conversing by a riverbank. Bear One stands upright on the upper-left, wearing a puzzled yet amused expression, holding an empty fishing net. Its oval speech bubble, outlined in smooth black, contains the text ``Did all the fish plan a vacation without telling us?'' in friendly, rounded hand-lettering. Bear Two sits on a rock at the lower-right, laughing while holding a tangled fishing rod caught in a tree branch above. Its rectangular speech bubble with rounded corners overlaps Bear One's, featuring the text ``Maybe they've finally learned about our sneaky fishing tricks!'' in playful italicized typography. The scene includes a lush forest with towering trees, mossy rocks, blooming flowers, and dynamic fish leaping from the river. The background is rich with color, with the bears' exaggerated expressions and speech bubbles dominating the composition. The comic style emphasizes bold outlines, dynamic angles, and whimsical details to highlight the humorous interaction.}{3pt}
\ltrow{151}{A high-resolution, sharply focused display case at a marine biodiversity museum showcases a lifelike great white shark replica positioned centrally. The shark, meticulously detailed with textured skin and realistic fin movement, is illuminated by soft, even lighting. Directly beneath it, a bold, high-contrast label reads ``Great White Shark (Carcharodon carcharias)'' in crisp, sans-serif typography. Below this, smaller explanatory text states ``Length: up to 6 meters'' and ``Habitat: Coastal waters worldwide,'' all perfectly aligned and undistorted. To the left, a series of neatly arranged informational panels feature concise, legible labels: ``Leatherback Sea Turtle - Largest Sea Turtle,'' ``Pacific Giant Octopus,'' and ``Lion's Mane Jellyfish,'' each with clear, flat typography. The background is dimly lit and out-of-focus, emphasizing the textual clarity and biological details. The entire exhibit is grounded in a neutral, uncluttered environment with sharp focus on the illuminated models and their precise, informative labels.}{3pt}
\ltrow{597}{A vibrant, humorous cartoon-style illustration of two friendly snails engaged in a playful conversation within a whimsical garden. The scene features oversized mushrooms, dew-covered leaves, colorful flowers, and tiny insects. In the upper-left corner, Snail One, with wide, exaggerated eyes and an exasperated expression, speaks from an oval speech bubble with soft, rounded edges and a thin, playful outline. The bubble contains the text ``You mean we moved only two inches since yesterday morning?'' in a friendly hand-drawn font, high-contrast and perfectly legible. In the lower-right corner, Snail Two lounges atop a small mushroom, smiling calmly, with a rectangular speech bubble featuring softly rounded corners and a slight upward tilt toward Snail One's bubble. The text ``Relax, friend! We're snails, speed isn't exactly our thing!'' appears in whimsical italic letters. The two bubbles overlap slightly at the bottom, creating a natural diagonal flow from top-left to bottom-right. The garden background is lush and detailed but softly blurred to prioritize the snails and text, rendered in a cohesive, stylized cartoon aesthetic with bold outlines and exaggerated expressions.}{0pt}
\end{minipage}
\caption{\textbf{Text rendering on LongText-Bench prompts (1024\texttimes1024).} Both DWD and AsymFlow are finetuned from FLUX.2 klein Base 9B, while L2P is finetuned from Z-Image Turbo. All images use the same seed. DWD preserves its base model's text rendering capability, whereas AsymFlow and L2P distort letters and words.}
\label{fig:app_longtext}
\end{figure}

AsymFlow~\citep{chen2026asymflow} and L2P~\citep{chen2026l2p} also remove the decoder, by adapting a pretrained latent model into a pixel model, which changes its sampling trajectory. We compare the official AsymFlow and L2P models against their respective base models, FLUX.2 klein Base 9B~\citep{flux2} and Z-Image Turbo~\citep{zimage}. Both methods are substantially worse than their base models on the reported metrics (Table~\ref{tab:app_asymflow}), especially in text rendering (Fig.~\ref{fig:app_longtext}). For AsymFlow, this is partly due to its sensitivity to the training data, which contains little text rendering, and partly due to the difficulty of adapting a latent trajectory to a pixel trajectory. In contrast, DWD matches the base model while preserving seed-level content.

DWD is also much cheaper to train, requiring only about a third of AsymFlow's training cost (350 vs.\ 1100 H100 GPU hours) even with a longer, saturated schedule (26k vs.\ 15k iterations). This is because DWD finetunes only the last few blocks with shallow backpropagation.

\begin{table}[H]
\centering
\caption{\textbf{Comparison with latent-to-pixel adaptation (1024\texttimes1024).} Z-Image Turbo results are from its official report~\citep{zimage}.}
\label{tab:app_asymflow}
\setlength{\tabcolsep}{0.5em}
\scalebox{0.9}{%
\begin{tabular}{@{}lccccc@{}}
\toprule
\textbf{Method} & \textbf{FID}\textdownarrow & \textbf{pFID}\textdownarrow & \textbf{OneIG}\textuparrow & \textbf{LongText}\textuparrow & \makecell[c]{\textbf{Training cost} \\ (H100 GPU hours)} \\
\midrule
FLUX.2 klein Base 9B & 15.13 & 18.69 & \textbf{0.556} & \textbf{0.909} & -- \\
\quad AsymFlow & 18.10 & 22.62 & 0.506 & 0.777 & 1100 (15k iters) \\
\rowcolor{blue!6} \multicolumn{1}{@{}>{\columncolor{blue!6}[0pt][\tabcolsep]}l}{\quad\textbf{DWD}} & \textbf{15.05} & \textbf{18.38} & \textbf{0.556} & 0.908 & \multicolumn{1}{>{\columncolor{blue!6}[\tabcolsep][0pt]}c@{}}{\hphantom{0}350 (26k iters)} \\
\midrule
Z-Image Turbo & -- & -- & \textbf{0.528} & \textbf{0.917} & -- \\
\quad L2P & -- & -- & 0.499 & 0.787 & -- \\
\bottomrule
\end{tabular}
}
\end{table}

\ifarxiv
\subsection{Preliminary Results on MiniMax-H3}
\label{app:mmh3}
We also apply DWD to 2-step and 4-step distilled MiniMax-H3 video models~\citep{chen2026pih3}. These models use a decoder distillation variant, where the final step's $\ell_1$ and perceptual target becomes the output of the reference model's native final step followed by the native decoder.

Due to the extreme compression of the MiniMax-H3 autoencoder, each token covers a 32\texttimes32 spatial patch over a maximum of 4 frames, so a linear head would have up to 4\texttimes32\texttimes32\texttimes3 = 12288 output dimensions, more than twice the DiT's hidden dimension of 5376. In addition, MiniMax-H3 is a single-stream model, whose transformer weights are shared across text, visual, and audio tokens. Therefore, a linear head has very limited capacity on these models, which benefit from a lightweight refiner (e.g., the DWD-conv head in Appendix~\ref{app:head}).

Please refer to our blog post~\citep{chen2026pih3} for video results. These results are purely preliminary, since we have not yet determined the optimal refiner architecture. 
\fi

\clearpage
\section{Additional Qualitative Results}
\label{app:qualitative}

\begin{figure}[H]
\centering
\begin{minipage}{\linewidth}
\qualhead
\qualpair{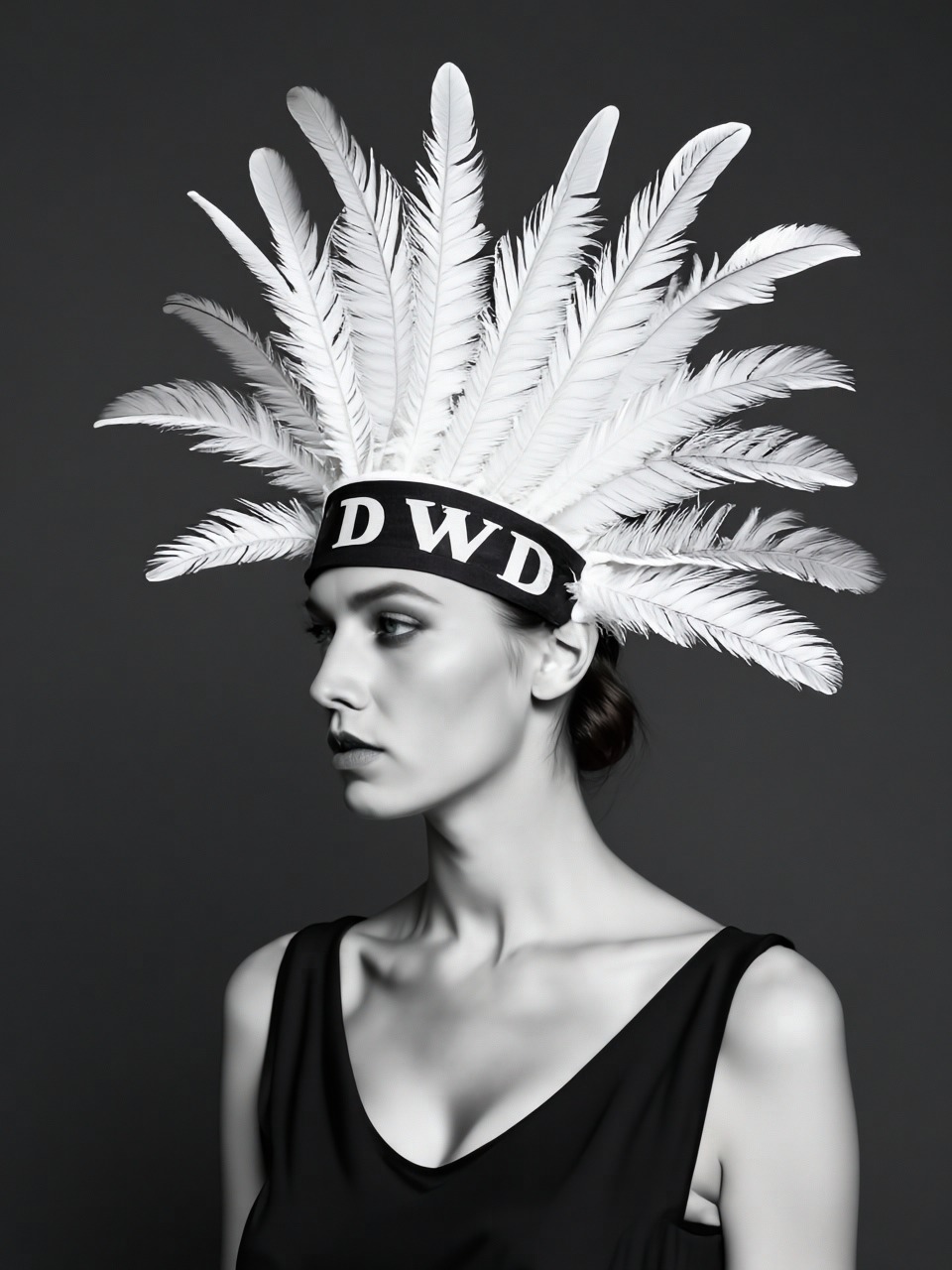}{Avant-garde fashion portrait, a feathered headdress with a band reading ``DWD'', stark studio lighting, black background.}{6pt}
\qualpair{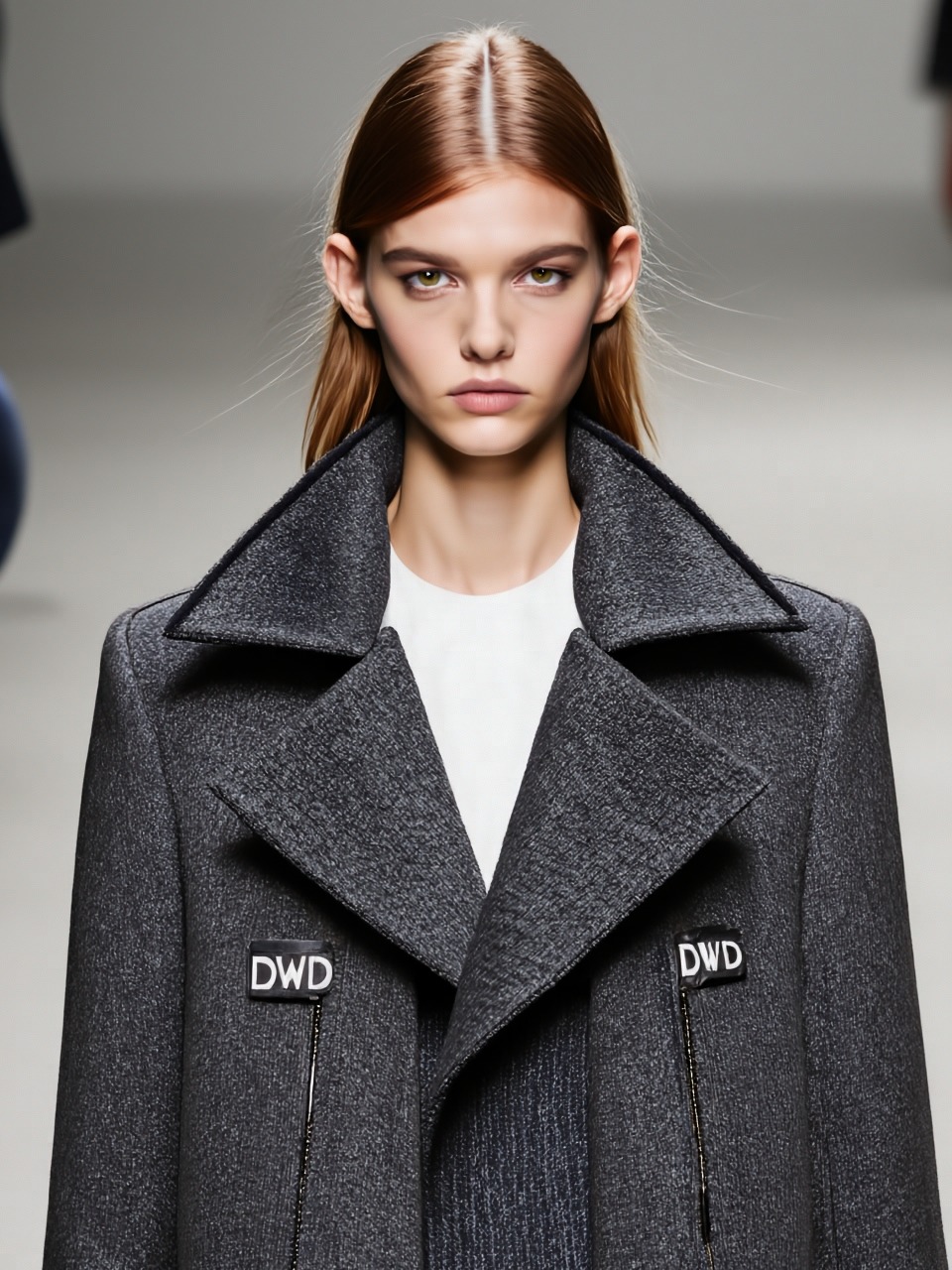}{A fashion photo of a model in a coat with ``DWD'' on the collar.}{0pt}
\end{minipage}
\caption{\textbf{Additional same-seed comparisons on AnisoFlow-L/16-T2I (1/3).} Feather barbs and wool weave are recovered by DWD; the baseline blurs both into flat regions.}
\label{fig:app_qual}
\end{figure}

\begin{figure}[H]
\centering
\begin{minipage}{\linewidth}
\qualhead
\qualpair{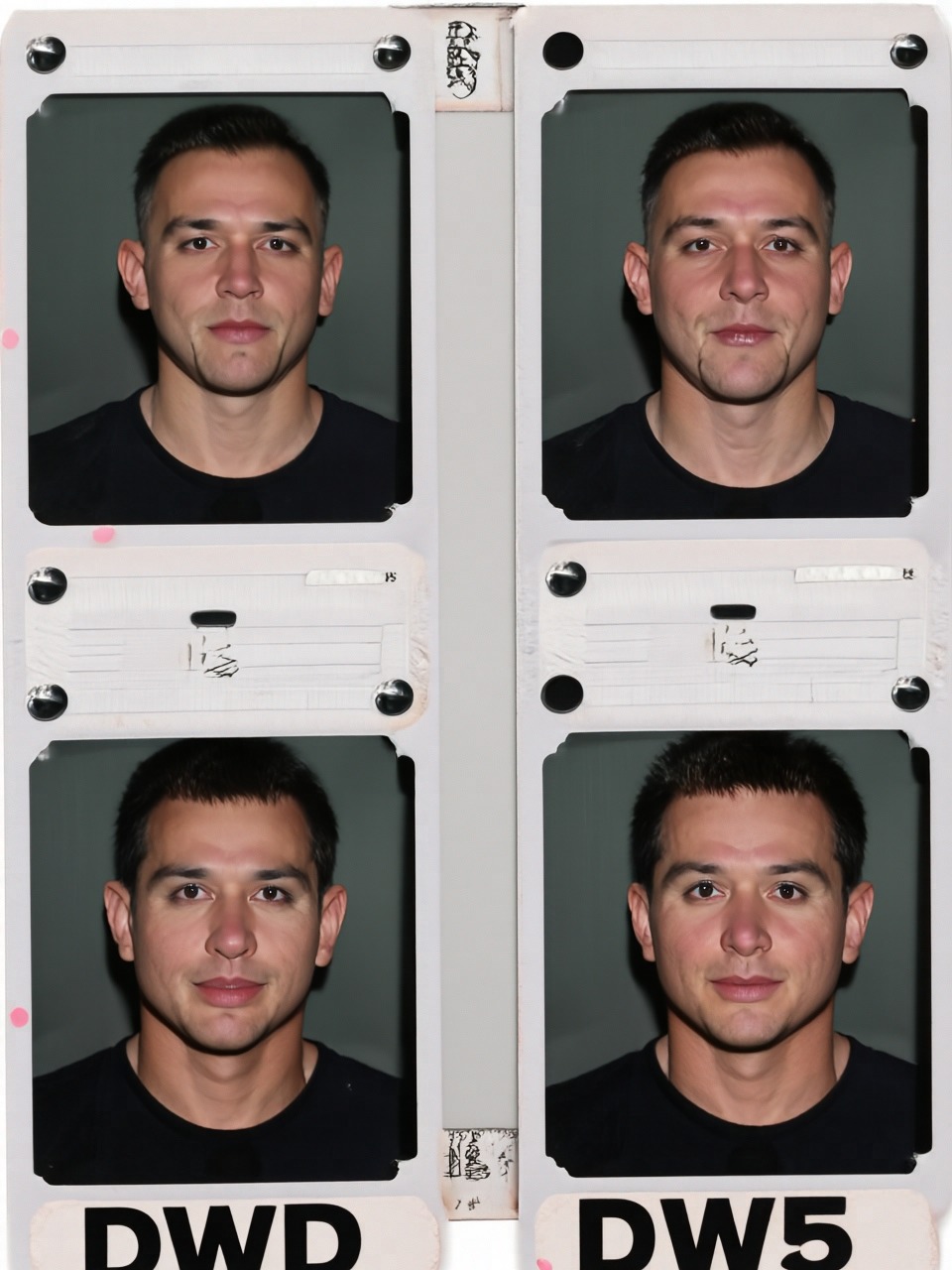}{A 1995 photo booth strip of four frames, ``DWD'' stamped across the bottom edge.}{6pt}
\qualpair{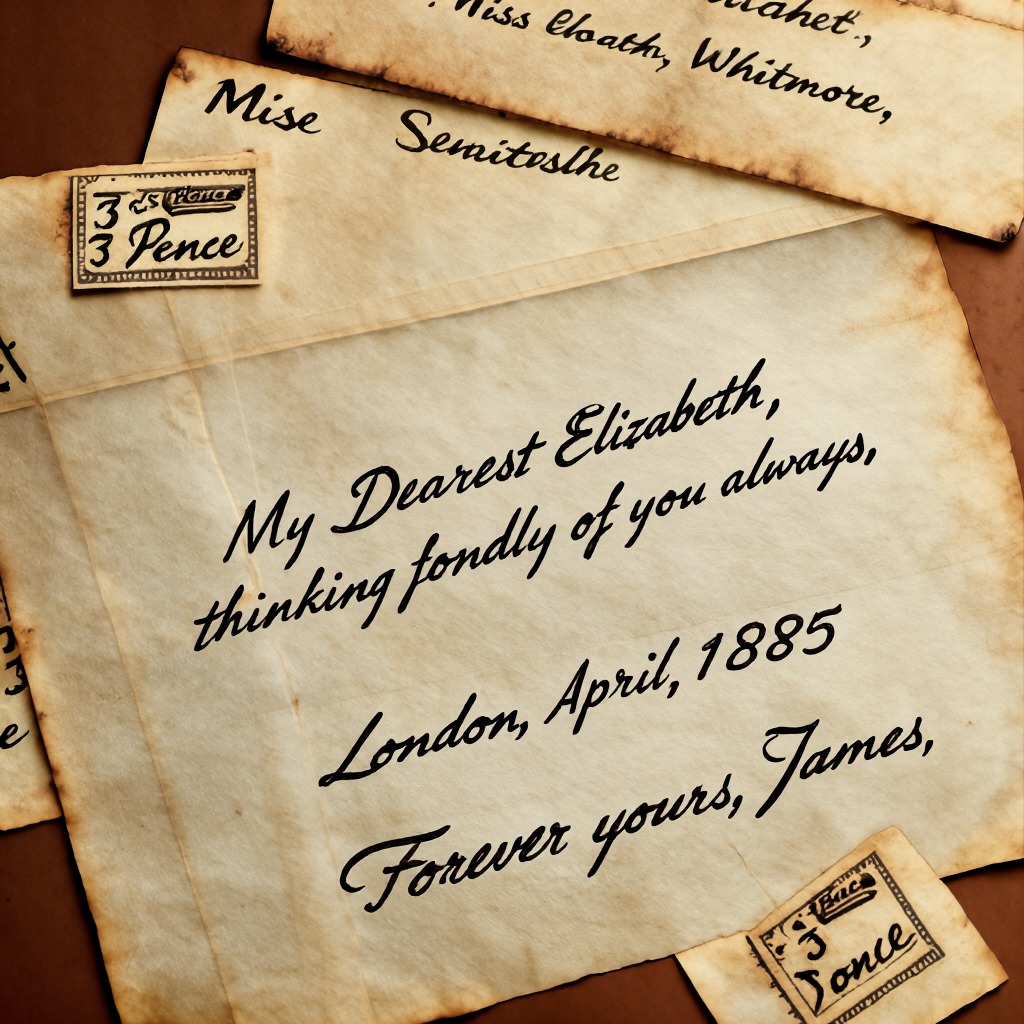}{A vintage letter rests on a slightly yellowed, textured parchment sheet, visibly folded with faint, delicate crease lines. The central text is penned in clear, elegant cursive: ``My Dearest Elizabeth, thinking fondly of you always,'' perfectly centered and legible. Below, smaller yet distinct script reads ``London, April 12th, 1885,'' in refined, evenly spaced handwriting. At the lower-right corner, soft cursive states ``Forever yours, James,'' subtly integrated into the design. Around the letter, vintage postal stamps with ``3 Pence'' inscriptions are delicately placed. Nearby, aged envelopes bear handwritten addresses: ``Miss Elizabeth Whitmore, Oxfordshire.'' Delicate stationery notes faintly labeled ``Cherished Sentiments'' add to the scene. All text is authentically handwritten, with thoughtful spacing and refined aesthetics. The atmosphere is bathed in warm, soft lighting, emphasizing the aged parchment, inked details, and nostalgic romanticism of a historical postal setting.}{0pt}
\end{minipage}
\caption{\textbf{Additional same-seed comparisons on AnisoFlow-L/16-T2I (2/3).} Skin texture in the photo booth frames and the small stamp and envelope lettering on the letter are sharper under DWD. The letter prompt is from LongText-Bench.}
\label{fig:app_qual2}
\end{figure}

\begin{figure}[H]
\centering
\begin{minipage}{0.85\linewidth}
\qualhead
\qualpair{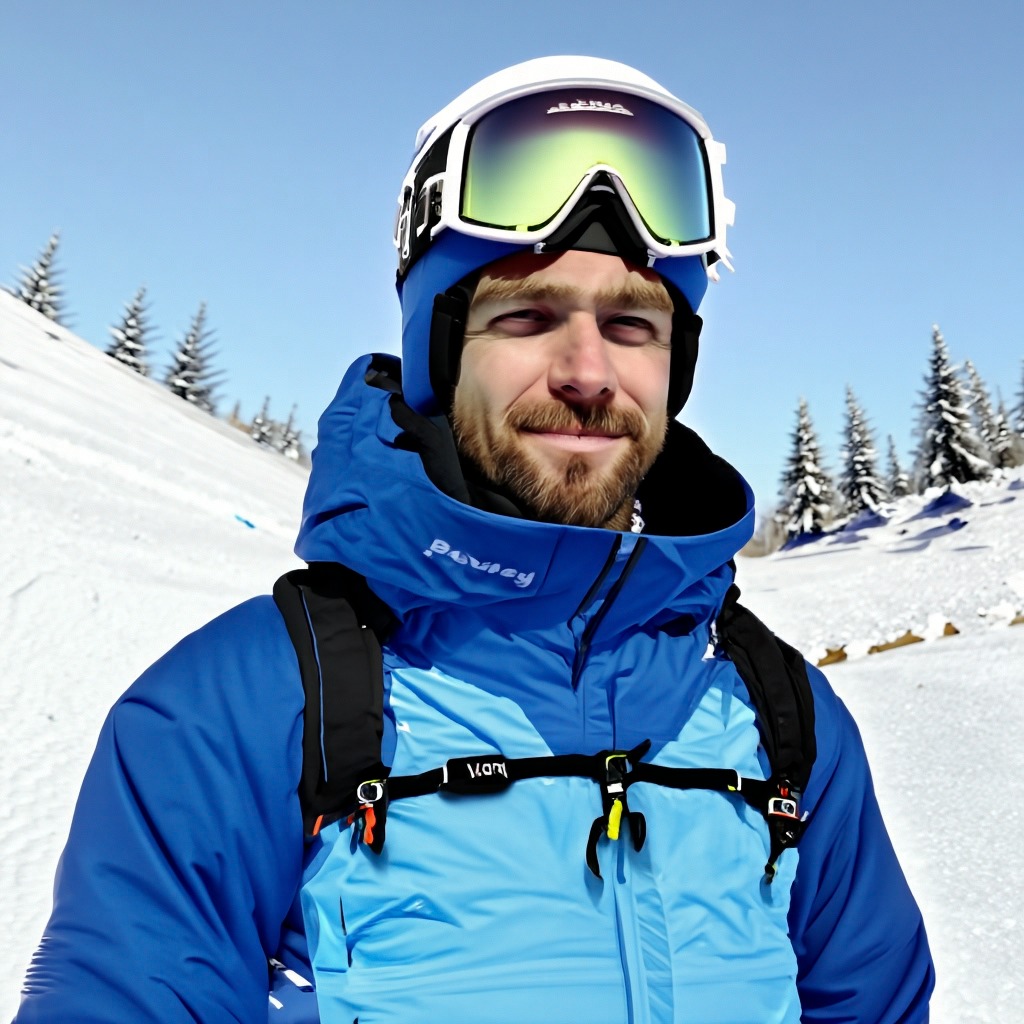}{A man in his skiing gear is posing for the camera.}{5pt}
\qualpair{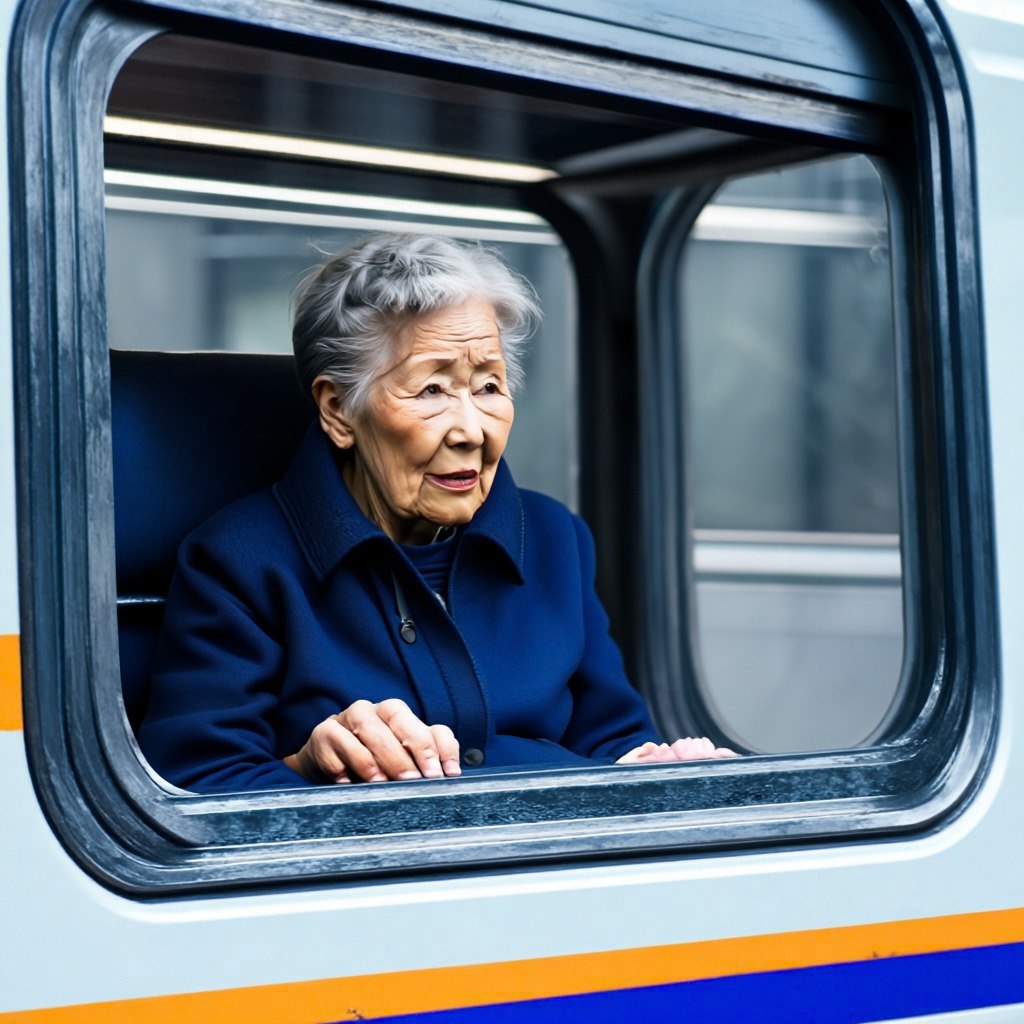}{An older woman riding a train while sitting under it's window.}{5pt}
\qualpair{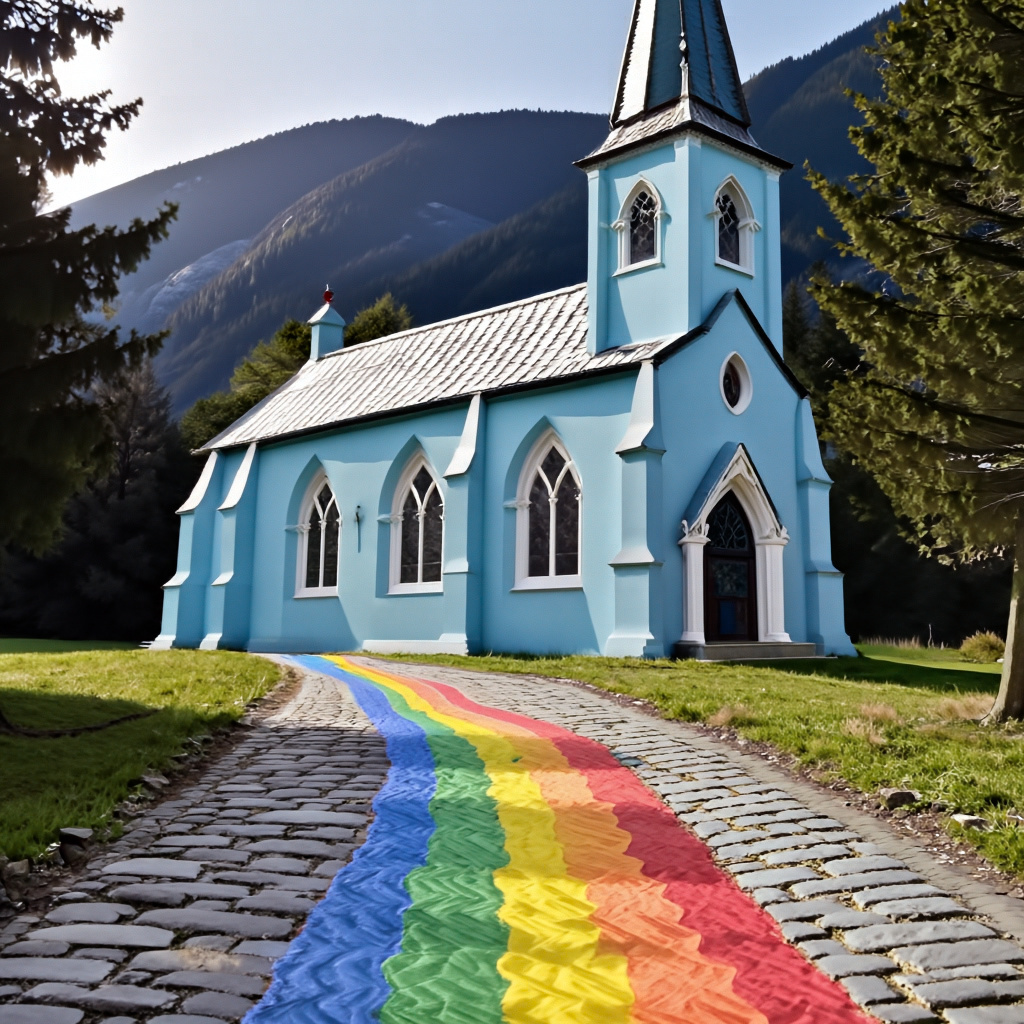}{A cobblestone trail with rainbow chalk stripes leads to a pale blue gothic church featuring white-framed arc windows and a single tower, surrounded by dark green trees and a sunlit Nordic mountain.}{0pt}
\end{minipage}
\caption{\textbf{Additional same-seed comparisons on AnisoFlow-L/16-T2I (3/3).} COCO and OneIG prompts. DWD recovers facial detail, fabric texture, and the individual cobblestones that the baseline smooths.}
\label{fig:app_qual3}
\end{figure}

\end{document}